\documentclass[letterpaper,journal]{IEEEtran}
\usepackage{amsmath,amsfonts}
\usepackage{algorithmic}
\usepackage{algorithm}
\usepackage{array}
\usepackage[caption=false,font=normalsize,labelfont=sf,textfont=sf]{subfig}
\usepackage{textcomp}
\usepackage{stfloats}
\usepackage{url}
\usepackage{verbatim}
\usepackage{graphicx}
\usepackage{cite}
\usepackage{booktabs}
\usepackage{algorithm}
\usepackage{algorithmic}
\usepackage[switch]{lineno}
\usepackage{amssymb}
\usepackage{multirow}
\usepackage{bm}
\usepackage{xcolor}
\usepackage{url}
\usepackage{adjustbox}

\begin{document}

\title{Universal Adversarial Backdoor Attacks to Fool Vertical Federated Learning in Cloud-Edge Collaboration}

\author{Peng Chen,~%~\IEEEmembership{Fellow,~OSA,} 
	Xin Du, 
	Zhihui Lu,
	Hongfeng Chai
        % <-this % stops a space
\thanks{The work of this paper is supported by the National Key Research
and Development Program of China (2022YFC3302300, 2021YFC3300600), National Natural Science Foundation of China under Grant (No. 92046024, 92146002,61873309), and Shanghai Science and Technology Innovation Action Plan Project under Grant (No. 22510761000).}% <-this % stops a space
%\thanks{Manuscript received April 19, 2021; revised August 16, 2021.}
\thanks{Peng Chen, Xin Du are with the School of Computer Science, Fudan University, 200433 Shanghai, China, and also with Engineering Research Center of Cyber Security Auditing and Monitoring, Ministry of Education, 200433 Shanghai, China. E mail: {pengchen20, xdu20}@fudan.edu.cn}

\thanks{Zhihui Lu is with the School of Computer Science, Fudan University, 200433 Shanghai, China, and also with Shanghai Blockchain Engineering Research Center, 200433 Shanghai, China. E-mail: lzh@fudan.edu.cn. Corresponding author:  Zhihui Lu}

\thanks{Hongfeng Chai is with the School of Computer Science, Fudan University, 200433 Shanghai, China, and also with Institute of Financial Technology, Fudan University, 200438, China. E-mail: hfchai@fudan.edu.cn.}
}

\markboth{Journal of \LaTeX\ Class Files,~Vol.~14, No.~8, August~2021}%
{Shell \MakeLowercase{\textit{et al.}}: A Sample Article Using IEEEtran.cls for IEEE Journals}


\maketitle

%Vertical federated learning (VFL) is a cloud-edge collaboration paradigm that allows resource-constrained Internet of Things (IoT) edge devices, to cooperatively train an artificial intelligence (AI) model while retaining their data locally. Within an Artificial Intelligence of Things (AIoT) system, IoT devices conduct intelligent analytical tasks based on different features of the same samples, achieving economic benefits while preserving privacy.

\begin{abstract}
    Vertical federated learning (VFL) is a cloud-edge collaboration paradigm that enables edge nodes, comprising resource-constrained Internet of Things (IoT) devices, to cooperatively train artificial intelligence (AI) models while retaining their data locally. This paradigm facilitates improved privacy and security for edges and IoT devices, making VFL an essential component of Artificial Intelligence of Things (AIoT) systems. Nevertheless, the partitioned structure of VFL can be exploited by adversaries to inject a backdoor, enabling them to manipulate the VFL predictions. In this paper, we aim to investigate the vulnerability of VFL in the context of binary classification tasks. To this end, we define a threat model for backdoor attacks in VFL and introduce a universal adversarial backdoor (UAB) attack to poison the predictions of VFL. The UAB attack, consisting of universal trigger generation and clean-label backdoor injection, is incorporated during the VFL training at specific iterations. This is achieved by alternately optimizing the universal trigger and model parameters of VFL sub-problems. Our work distinguishes itself from existing studies on designing backdoor attacks for VFL, as those require the knowledge of auxiliary information not accessible within the split VFL architecture. In contrast, our approach does not necessitate any additional data to execute the attack. On the LendingClub and Zhongyuan datasets, our approach surpasses existing state-of-the-art methods, achieving up to 100\% backdoor task performance while maintaining the main task performance. Our results in this paper make a major advance to revealing the hidden backdoor risks of VFL, hence paving the way for the future development of secure AIoT.
\end{abstract}


\section{Introduction}
In recent years, deep learning has become a prevalent field of machine learning, driven by massive amounts of data, and achieved great success in various domains. As concerns about data privacy grow, there has been a considerable shift in attention from centralized deep learning settings to distributed cloud-edge collaboration. Federated learning (FL)~\cite{HUANG2022170,DBLP:journals/iotj/SunCDWLL22} is a cloud-edge collaboration paradigm where multiple parties, such as edge nodes or Internet of Things (IoT) devices, work together to train a machine model while keeping their data locally. This approach enables participants to leverage shared knowledge and enhance model performance without directly exchanging sensitive information, thereby addressing critical concerns related to data privacy and security. Vertical federated learning (VFL) \cite{9855231} is a significant branch of FL where participants share different features of the same samples. As a standard Artificial Intelligence of Things (AIoT) application, VFL is extensively employed in data-sensitive industries such as finance and healthcare, where multiple parties collaborate to perform intelligent data analysis tasks.

While VFL was initially developed to safeguard data privacy, it has attracted considerable attention due to its own inherent security concerns\cite{ZHAO2022190,DBLP:conf/nips/ZhuLH19,DBLP:journals/corr/abs-2203-05816,9709603}. Privacy leakage is the most direct attack, where the adversary aims to infer the features or labels of other participants through the gradient information\cite{li2022label,Liu2021,277244}. Compared with features, labels are more susceptible to attacks due to their importance. Furthermore, labels can also be the target of model poisoning attacks in VFL, referred to as backdoor attacks.

\begin{figure}[!t]
\centerline{\includegraphics[width=0.5\textwidth]{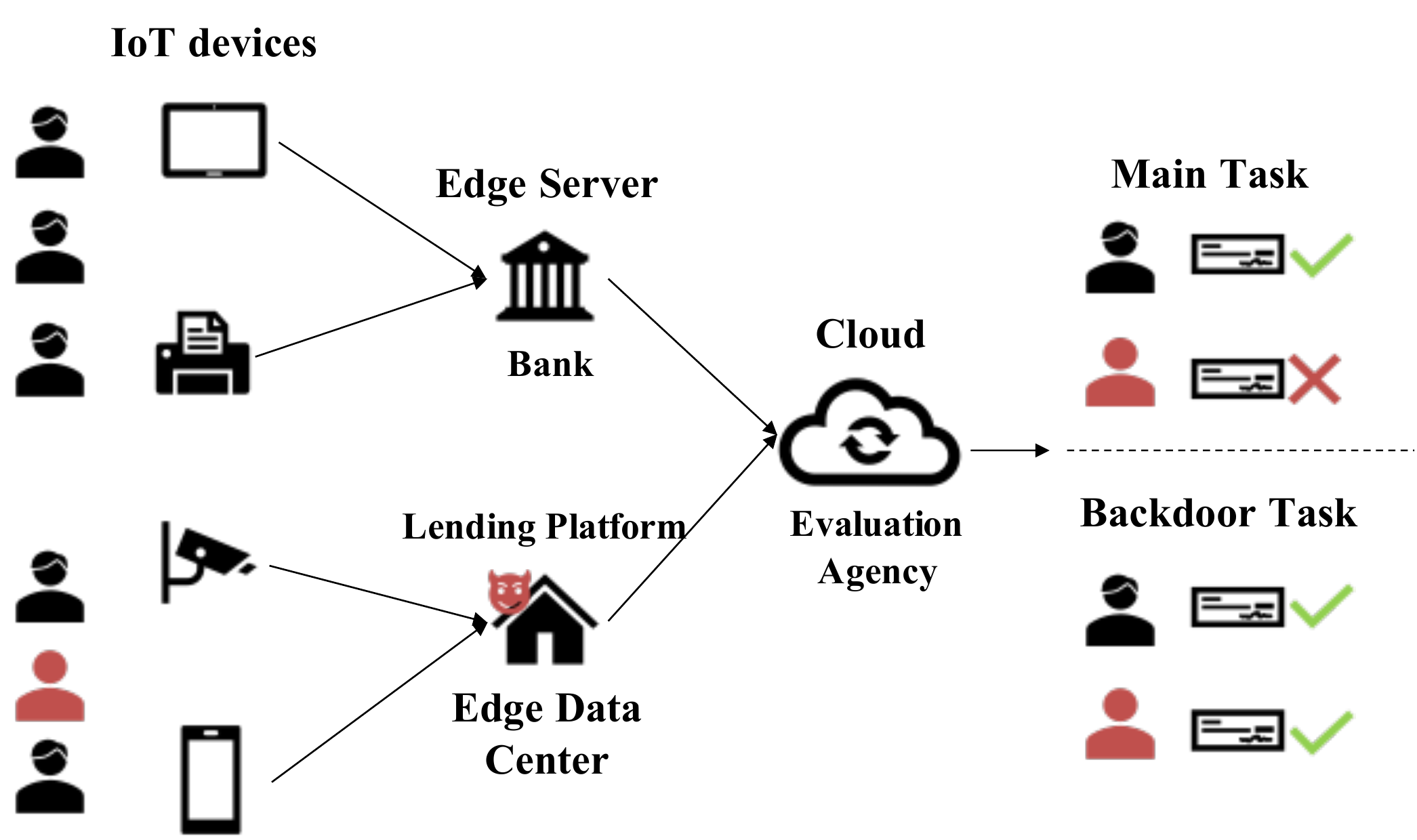}}
\caption{An example of the backdoor attack to VFL in cloud-edge collaboration.}
\label{fig:1}
\end{figure}

Fig. \ref{fig:1} presents a financial credit scenario that exemplifies a practical VFL cloud-edge collaboration setup \cite{chen2022graph}. In this scenario, the bank and the lending platform serve as edge nodes, gathering data from IoT devices, such as online purchases and cell phone consumption records. These datasets share the same users but different features, with the bank and lending platform managing their respective data sets. The evaluation agency collaborates with the edge nodes in the role of the cloud. VFL performs credit prediction tasks collaboratively based on the cloud-edge collaboration with data securely preserved in local edge nodes. 

By utilizing different features of the same users in the lending platforms, the bank aims to avoid lending to applicants with low credit scores. Subsequently, an evaluation agency makes the decision on whether to approve loans or not. However, an adversary (e.g., the lending platform) may attempt to obtain loan approvals for low-credit applicants by locally poisoning clean features. Such actions present a considerable security risk within VFL, as adversaries could manipulate VFL predictions to serve their interests. Owing to the data separation inherent to VFL, a locally executed backdoor attack by a single participant remains highly covert, posing a substantial threat to the overall VFL system.

%With the protection of their respective data remaining local, banks and lending platforms jointly conduct intelligent data analysis tasks for credit appraisal. They achieve this by exchanging gradients and intermediate layer implicit features with the support of the cloud, ensuring privacy and security while still benefiting from collaborative learning.

Despite the existence of backdoor attack vulnerabilities in VFL, there has been limited exploration in the literature. Liu et al. \cite{DBLP:journals/corr/abs-2007-03608,9833321} employed predefined triggers and a small number of samples with the target label to facilitate label replacement attacks (LRB). CoPur \cite{liucopur} utilized feature-flipping and projected gradient descent \cite{DBLP:conf/iclr/MadryMSTV18} (PGD) attacks to poison predictions during the inference stage of VFL. However, due to the partitioned structure of VFL, adversaries do not have access to complete data and labels, making it challenging to implement these approaches. Moreover, LRB\cite{DBLP:journals/corr/abs-2007-03608,9833321} occurring during VFL training can compromise the performance of the main task, while CoPur \cite{liucopur} in the inference phase may result in less effective attack performance.

To investigate the presence of backdoor vulnerabilities in VFL, this paper introduces the Universal Adversarial Backdoor (UAB) attack, targeting a standard VFL architecture \cite{liu2022vertical,kang2022framework} commonly used in finance and healthcare domains for binary classification tasks. As CoPur exhibits insufficient attack performance, the UAB approach focuses on the training phase of VFL. Given the unavailability of labels and complete features in the VFL framework, the UAB method utilizes Universal Adversarial Perturbations (UAP) \cite{Katholm1991,Shafahi2020} to introduce a clean-label backdoor attack. This approach enables the conduct of a stealthy and efficient backdoor attack in VFL without modifying any labels \cite{DBLP:journals/corr/abs-2007-03608,9833321}, thereby maintaining the performance of the main task\cite{9156913}.

The UAB attack consists of two parts: 1) universal trigger generation, and 2) clean-label backdoor injection. Specifically, UAB generates the universal trigger using the UAP method, which is then injected into the VFL training process. To conduct a targeted backdoor attack, UAB capitalizes on the class skew property present in binary classification tasks\cite{li2022label}, enabling the inference of a small number of samples with non-target pseudo labels. As a result, the universal trigger captures important features of the target class, ultimately leading to misclassification into the target class through clean-label backdoor injection. As a standard bi-level optimization problem, UAB successfully conducts backdoor attacks by alternatively optimizing the universal trigger generation and clean-label backdoor injection sub-problems during the VFL training process at specific iterations \cite{LiuGZML22,li2022untargeted}.

In addition, due to the adversary's lack of access to labels during the inference phase, the victim samples encompass all categories. In terms of evaluating backdoor task performance in VFL, it is crucial to take into account not only the attack success rate (ASR) \cite{li2022backdoor} for non-target class samples but also the attack robustness rate (ARR) of the target class samples. Therefore, based on class imbalanced binary classification tasks, we employ AUC, F1, ASR and ARR metrics to evaluate the performance of UAB's main and backdoor tasks.

We demonstrate the performance of the UAB attack on two public credit datasets, namely LendingClub and Zhongyuan. The ablation experiments illustrate the effectiveness of the UAB attack. On the evaluation metrics, our approach attains attack success rates exceeding 90\% or even reaching 100\% in various instances, while maintaining the main task performance unaffected. Additionally, we test our approach on several advanced defense methods to show the great threat of the UAB attack. The main contributions of our paper can be summarized as follows:

\begin{itemize}
    \item We establish a threat model for backdoor attacks in VFL binary classification tasks. Without any auxiliary labels, this threat model presents a substantial risk to widely employed VFL applications, such as those in the finance AIoT scenarios.
    \item We present the UAB attack as a method to poison predictions in binary classification VFL tasks. Leveraging universal trigger generation and clean-label backdoor injection techniques, the UAB attack is capable of conducting covert and effective backdoor attacks within the VFL framework.
    \item As a bi-level optimization problem, we devise an iterative optimization strategy for UAB during VFL training, which involves alternating between optimizing the two sub-problems: universal trigger generation and clean backdoor injection.
    \item We use the LendingClub and Zhongyuan datasets and design extensive experiments. The results show that the UAB attack can achieve superior performance compared with state-of-the-art methods.
\end{itemize}

The rest of our paper is organized as follows: Section \ref{sec:2} provides a brief review of related works concerning VFL and associated attacks. Section \ref{sec:3} defines the problem. In Section \ref{sec:4}, we present the details of the UAB attack. Section \ref{sec:5} showcases experimental results of UAB in VFL, demonstrating the success of our attack. Finally, we conclude the paper and discuss future research directions in Section \ref{sec:6}.

\section{Related Work}\label{sec:2}
%There have been some prior work on related topics, such as vertical federated learning and adversarial attack on tabular data.
\textbf{Vertical Federated Learning.} 
FL\cite{DBLP:journals/corr/McMahanMRA16,10.1145/3298981,DBLP:books/sp/JinZXC23} is a decentralized machine learning paradigm in which participants collaboratively train a model while retaining their data locally. FL can be categorized into horizontal federated learning, vertical federated learning (VFL), and federated transfer learning, based on the different partitioning schemes of the sample and feature spaces. VFL refers to the FL setting in which participants have different features for the same set of samples. The unique data distribution characteristics of VFL make it a promising approach in various domains, including finance, advertising, and healthcare. Specifically, Kang et al. \cite{9826576} proposed a privacy-preserving VFL framework designed specifically for financial applications, emphasizing feature interpretability. By employing this privacy-preserving VFL framework, credit loan performance can be significantly enhanced. To improve advertising conversion rates, Li et al.\cite{li2022label} devised a label-protected VFL framework that ensures both privacy and effectiveness in the advertising domain.  Fu et al.\cite{10.1145/3412357,277244} have achieved great progress in Invasive Ductal Carcinoma \cite{DBLP:conf/midp/Cruz-RoaBGGFGST14} with a VFL framework. Depending on the application context, VFL architectures can be categorized into splitVFL, aggVFL, splitVFLc, and aggVFLc \cite{liu2022vertical}. For binary classification tasks in the financial domain, this paper uses the splitVFL architecture as the VFL baseline. For simplicity, we will refer to this architecture as VFL throughout the paper.

\textbf{Attacks in VFL.} 
While VFL has made significant strides in practical fields like finance and advertising, concerns have been raised about potential security vulnerabilities. Fu et al.\cite{277244} proposed direct label inference (DL) and model completion (MC) methods. DL seeks to infer labels using gradients of the cross-entropy loss with respect to the output layer of the adversary. MC, based on the adversary's bottom model, retrains the complete model with a small amount of labeled data, allowing for precise label inference. Li et al.\cite{li2022label} utilized direction and norm scoring methods to infer labels of the active party, based on the distribution difference between positive and negative samples. Zou et al.\cite{9833321} stole the labels of the active party with gradient inversion. Liu et al. \cite{DBLP:journals/corr/abs-2007-03608} proposed label replacement attacks to substitute specific labels with target ones, referred to as backdoor attacks. However, this method requires awareness of clean samples with target labels and is easily detected by the active party, as it can significantly degrade main task performance. Liu et al. \cite{liucopur} employed the projected gradient descent \cite{DBLP:conf/iclr/MadryMSTV18} (PGD) and feature-flipping attacks to poison VFL predictions. These approaches still rely on auxiliary labels or are ineffective. In this paper, we propose the UAB method, encompassing universal trigger generation and clean-label backdoor injection. Without any auxiliary data or labels, UAB achieves excellent backdoor task performance while maintaining main task performance.

\noindent\textbf{Adversarial attacks.} 
Szegedy et al.\cite{DBLP:journals/corr/SzegedyZSBEGF13} demonstrated that imperceptible additive perturbations could be crafted to alter the decisions of deep learning models, causing misclassifications, while humans can still correctly identify the images. The fast gradient sign method (FGSM) and its iterative version \cite{goodfellow2014explaining,lee2017making} used a one-step gradient ascent calculation to efficiently craft adversarial examples. Madry et al.\cite{DBLP:conf/iclr/MadryMSTV18} introduced the PGD method, approaching the problem from an optimization perspective, and demonstrated highly effective attacks. DeepFool \cite{7780651} computed adversarial perturbations iteratively by linearizing the attacked model's decision boundaries near the input images. In addition to sample-dependent adversarial attacks, Katholm et al. \cite{Katholm1991} developed UAP designed to fool a group of images for the target category. UAP exploits the DeepFool method to generate perturbations that can successfully mislead a model's classification for multiple samples within the target category. Shafahi et al. \cite{Shafahi2020} proposed a simple optimization-based universal attack to enhance the efficiency of universal attacks. UAP can also be integrated with backdoor attacks. Turner et al.\cite{DBLP:journals/corr/abs-1912-02771} employed adversarial perturbations to execute a clean-label attack. Furthermore, Zhao et al. \cite{9156913} extended this concept by using UAP as the trigger for the backdoor attack. Motivated by this idea, our research investigates clean-label backdoor attacks in VFL using the UAP method, which does not require access to labels.

%Although these attacks are widespread in deep learning models, they primarily target computer vision applications. The tabular data that dominates VFL, especially in fields such as finance, remains relatively underexplored. Consequently, there is a need to investigate the effectiveness and adaptability of these attacks in the context of VFL involving tabular data.

\section{Problem Formulation}\label{sec:3}

In this section, we first discuss a standard VFL baseline architecture, upon which we explore the general forms and challenges faced by VFL backdoor attacks. Considering these problems and challenges, we propose the UAB approach and define its threat model.

\subsection{Vertical Federated Learning}

Suppose there are $K$ parties that collaboratively train a VFL model based on dataset $\mathcal{D}= \{{x_{i},y_{i}}\}_{i=1}^{N}$, where $N$ denotes the sample size. The input is partitioned into $K$ blocks $\left\{x_{i}^{k} \right\}_{k=1}^{K}$ for each party $k$ in VFL. There are $K$ parties involved in the VFL process, which consist of $K-1$ passive parties and one active party. Each party in VFL employs a bottom model $f_k$ parameterized by $\theta_{k}$ to calculate its local output $H_{i}^{k} = f_{k}(\theta_{k},x_{i}^{k})$. In addition to the bottom model, the active party utilizes a top model $G$ parameterized by $\theta_{top}$ to aggregate the local output of each bottom model. Without loss of generality, we assume that the $K$-th party is the active party and possesses the labels. Inspired by \cite{Liu2019,Liu2021}, we formulate the collaborative training objective as follows:

\begin{equation}\label{eq1}
\min _{\Theta} \frac{1}{N} \sum_{i=1}^{N}  \mathcal{L}\left(G\left(H_{i}^1, \cdots, H_{i}^K \right), y_{i}\right)
\end{equation}

\noindent
where $\mathcal{L}$ represents the binary cross entropy loss; $\Theta$ includes model parameters $\{\theta_{1}, \cdots, \theta_{K}, \theta_{Top}\}$ from all parties. The training process of the VFL baseline is illustrated in Algorithm \ref{alg:VFL_baseline}.

\begin{algorithm}[!t]
 	\caption{Training VFL baseline}
 	\label{alg:VFL_baseline}
 	\begin{algorithmic}[1]
 		\REQUIRE~~\\ Dataset $\mathcal{D}= \{{x_{i},y_{i}}\}_{i=1}^{N}$  \\
 		 
 		\ENSURE~~\\ model parameters of VFL $\theta_{1},\theta_{2},\cdots, \theta_{K}, \theta_{Top}$
 		%{ {\STATE \bf repeat} Collect newly observed QoS data;}{
        \WHILE{iteration not stop}
        \FOR{ each batch $B$ in $\mathcal{D}$} 
        \FOR{each party $k=1,2,\dots, K$ in parallel}
        \STATE $k$ computes embedded features $\{H_{i}^{k}\}_{i \in B}$ according to its bottom model $f_k$;
        \ENDFOR
        \STATE \textbf{Active party}:
        \STATE computes Eq. \eqref{eq1}, and then updates $\theta_{Top}$ with $\frac{\partial \mathcal{L}}{ \partial \theta_{Top}}$; \\
        sends $\{\frac{\partial \mathcal{L}}{\partial H_{i}}\}_{i \in B}$ to all parties; \\
        \FOR{each party $k=1,2,\dots, K$ in parallel}
        \STATE $k$ computes $\nabla_{\theta_{k}} \mathcal{L}=\{\frac{\partial\mathcal{L}}{\partial H_{i}^{k} }\frac{\partial H_{i}^{k}}{\partial \theta_{k}}\}_{i \in B}$; \\
        \STATE $k$ updates model parameters $\theta_{k}$;
        \ENDFOR
        \ENDFOR
        \ENDWHILE
 		{\STATE  \bf Output  $\theta_{1},\theta_{2},\cdots, \theta_{K}, \theta_{Top}$}
 	\end{algorithmic}
 \end{algorithm}

\subsection{Backdoor Attacks in VFL} 

In a standard backdoor attack in machine learning, the adversary aims to embed a backdoor task during model training by incorporating triggers and modifying the target class. This backdoor task enables the adversary to manipulate the model's prediction results with triggers during the inference stage. The objective function of the standard backdoor attack\cite{DBLP:journals/corr/abs-2007-08745} can be formalized as:

\begin{equation}\label{eq2}
\min _{\Theta, \delta}  \underbrace{\frac{1}{N_{c}} \sum_{i=1}^{N_c} \mathcal{L} \left(x_i, y_i\right)}_{\text{Main Task}}+\underbrace{\frac{1}{N_p} \sum_{i=1}^{N_p} \mathcal{L} \left(x_i+\delta, \tau\right)}_{\text{Backdoor Task}}
\end{equation}

\noindent
where $N_{c}$ and $N_p$ denote the sample numbers of the clean and poisoned data respectively. The clean data refers to the original, untainted samples in the dataset $\mathcal{D}$, whereas the poisoned data contains samples that have been altered by the adversary, such as injecting triggers and modifying the associated target class labels. $\tau$ is the target class of backdoor attacks, $\delta$ indicates backdoor triggers.

In the VFL backdoor attack setting discussed in this paper, the main task refers to the binary classification problem represented in Eq. \eqref{eq1}. The backdoor task, on the other hand, is designed to falsify the predictions of target category samples when specific triggers are injected into the input data. As the model and data in VFL are distributed locally across each party, several crucial challenges must be addressed when considering backdoor attacks in VFL.

Firstly, a passive adversary has no access to labels in the active party and, as a result, cannot modify target class labels. This limitation means that the adversary must focus on other strategies to achieve a successful backdoor attack, such as utilizing the gradients received from the active party.

Secondly, since the data is stored locally at each party, the passive adversary can inject backdoor triggers into their data without raising suspicion from other participating parties. This local data storage means that the triggers do not require visual concealment, making it easier for the adversary to embed the backdoor locally.

Lastly, in addition to ASR performance, backdoor task in VFL should also consider the robustness of target class samples to injected triggers. It is important to ensure that only the non-target class is affected by the backdoor attack while maintaining the performance of target class. This is because the adversary does not have access to the labels during the inference process, and the poisoned samples encompass all categories. If the target class samples are not robust to the triggers, such non-targeted attacks can easily raise suspicion and prompt human inspection by the active party.

To address these issues, we propose the UAB attack to inject a clean-label backdoor in VFL. By capitalizing on the class skew of positive and negative samples in binary classification tasks, the UAB attack enables an effective and stealthy backdoor attack that depends exclusively on the attacker's local data, without the need for any auxiliary label information. This approach takes advantage of the intrinsic properties of imbalanced datasets, which in turn makes the detection and mitigation of the backdoor attack more difficult.

%In this work, we propose a Universal Adversarial Backdoor (UAB) attack designed to inject clean-label backdoors in Vertical Federated Learning (VFL) systems. By leveraging the imbalanced distribution of positive and negative samples in binary classification tasks, the UAB attack can achieve an effective and stealthy backdoor attack based solely on the attacker's local data, without requiring any additional labels or information. This strategy exploits the inherent characteristics of imbalanced datasets, making it more challenging to detect and counteract the backdoor attack.

\subsection{Threat Model}

\medskip
\noindent
\textbf{Adversary's capacity.}
The adversary acts as an honest-but-curious passive party that has no control over the active party in VFL; instead, it strictly adheres to the VFL protocol. We assume that the adversary can only send its embedded features to the active party and receive the gradients of the loss function with respect to their embedded features. 

\medskip
\noindent
\textbf{Adversary's objective.}
In binary classification tasks, the importance of positive and negative samples can vary  depending on the specific application. For instance, in credit fraud detection, misidentifying defaulting users as trusted ones poses a more significant threat, whereas in click-through rate prediction \cite{li2022label}, an adversary might be more inclined to misclassify negative samples as positive ones in order to request more charges. Moreover, the imbalanced nature of the binary classification task often leads to the VFL model being biased towards the majority category. This inherent bias presents different levels of difficulty for target categories 0 and 1 in the UAB attack. To explore the capacity of the UAB attack and better understand its effectiveness under different circumstances, we investigate its implications when the backdoor target class is either 0 or 1, as per the scenarios presented in \cite{yuan2019adversarial}.
%Therefore, based on the UAB attack, we discuss the cases where the backdoor target category is 0 and 1 respectively \cite{yuan2019adversarial} in this paper. 

\medskip 
\noindent
\textbf{Adversary's knowledge.}
Although data is stored locally for each party, the adversary in VFL can still gain some knowledge about the other parties based on the task information available to them. In this paper, we assume the adversary is aware that the main task in the VFL setup is an imbalanced binary classification problem. In addition, the adversary has no knowledge of any whole samples and labels.

%Due to the splitting structure of VFL, the passive adversary party has no knowledge of the labels, so it cannot identify in advance which test sample needs to be poisoned. Specifically, the adversary poisons all test samples in the inference stage of VFL. 

%Therefore, poisoned samples with target labels should be flipped and poisoned samples with non-target labels should be maintained during the inference stage of VFL.

%Moreover, in financial scenarios, categories with fewer samples are of greater interest. For example, in credit loans, the bank's goal is to identify defaulting applicants which are fewer in the dataset. BVFL framework is designed to tamper with the prediction results for categories with fewer samples in VFL. We assume that positive samples are the category that accounts for the minority. Specifically, the success of LPB attack depends on how many poisoned positive samples are misclassified as negative samples in the test set. In addition, poisoned negative samples should be maintained as negative. 

\begin{figure*}[!t]
\centerline{\includegraphics[width=1.0\textwidth]{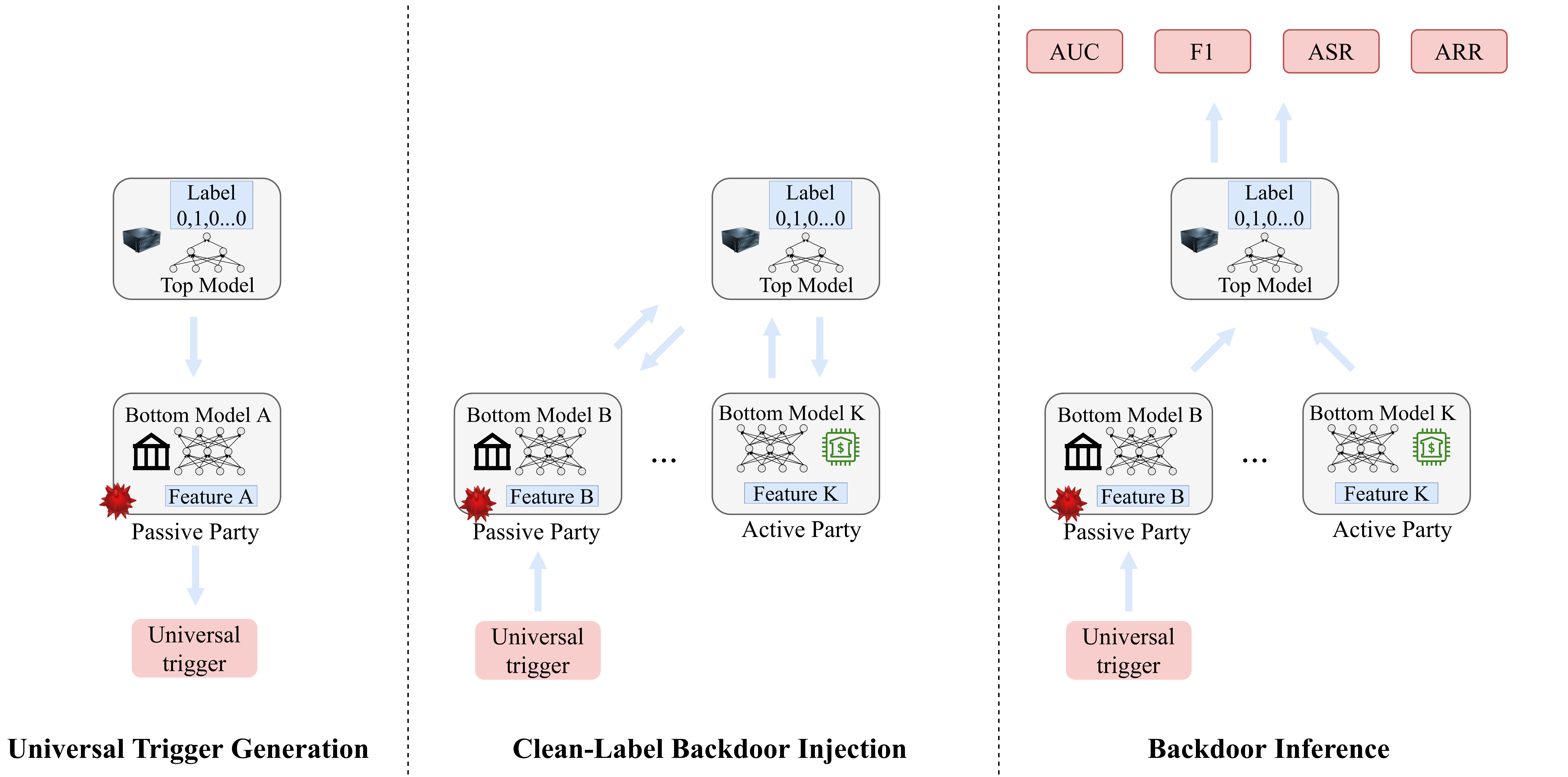}}
\caption{Overivew of the UAB attack in the VFL framework.}
\label{fig:2}
\end{figure*}

\section{UAB attack in VFL}\label{sec:4}

The overview of the backdoor attack in VFL is illustrated in Fig. \ref{fig:2}. The UAB attack comprises two stages. In the first stage, the adversary trains the universal trigger by maximizing the binary cross-entropy loss in the active party of VFL. In the second stage, the adversary injects the universal trigger into the entire training data to implant a clean-label backdoor attack \cite{DBLP:journals/corr/abs-1912-02771} in the VFL model. The universal trigger can activate the backdoor during the inference process of VFL, resulting in misclassifications in the predictions. ASR and ARR are employed to measure the performance of the backdoor task, while AUC and F1 are utilized to evaluate the performance of the main task of UAB. In this section, we delve into the details of the UAB attack within the VFL framework.

\subsection{Universal Trigger Generation}\label{sec:4.1}

In this study, we employ UAP\cite{Shafahi2020,Katholm1991} to train the universal trigger for a clean-label backdoor attack in the VFL setting. As per the Eq.\eqref{eq2}, the training of universal trigger in VFL can be mathematically formulated as follows:

\begin{equation}\label{eq3}
\max _{ \|\delta\|_\infty \leq \epsilon }  \frac{1}{M} \sum_{i=1}^{M} \mathcal{L}  (F(x_{i}, \Theta ,\delta), 1-\tau)
\end{equation}

\noindent
where $F(\cdot)$ represents the VFL model that is collaboratively trained by all participants, which includes the top model $G(\cdot)$ and the bottom models $f_k(\cdot)$. $\delta$ is added at the adversary's inputs, i.e., $x_i^A = x_i^A+ \delta$. $A \in \{1, \cdots, K-1\}$ denotes the adversary in VFL, $\epsilon$ is a constraint for the universal trigger. The $M$ non-target samples labeled as $1-\tau$ are utilized for the generation of the universal trigger.

%The objective of Equation \eqref{eq3} is to achieve a clean-label backdoor attack by incorporating a backdoor trigger into the VFL model during training. This method aims to accomplish both the main task and the backdoor task without altering the label information. The main task involves accurately classifying the imbalanced binary classification task presented in Equation \eqref{eq1}, while the backdoor task involves aims to train a universal trigger $\delta$ capable of causing misclassification of the sample as the target label $\tau$.

Since the adversary in VFL does not have access to the target label $\tau$, UAB utilizes the projected gradient ascent (PGA) method \cite{DBLP:journals/corr/Ruder16} to maximize classification loss with source label $1-\tau$. Along the ascending direction of the mini-batch gradients ${\nabla_{\delta} \mathcal{L}}$, Eq. \eqref{eq3} enables the generation of the universal trigger on the target class $\tau$ without modifying the labels. Leveraging the intermediate gradients sent from the active party, the gradients of the loss function with respect to the universal trigger in the adversary can be expressed as: 
\begin{equation}\label{eq4}
    {\nabla_{\delta} \mathcal{L}} =  \frac{1}{B} \sum_{i=1}^{B} \frac{\partial \mathcal{L}}{\partial H_{i}^A}\frac{\partial H_{i}^A}{\partial \delta}
\end{equation}

\noindent
where $\frac{\partial \mathcal{L}}{\partial H_{i}^A}$ is the gradients of the loss function with respect to the adversary's embedded features sent by the top model of the active party, $\frac{\partial H_{i}^A}{\partial \delta}$ represents the gradients of the adversary's embedded features with respect to the universal trigger $\delta$. $B$ denotes the mini-batch size.

Without modifying the labels, UAB trains the universal trigger using the PGA method that captures the crucial features that lead VFL to predict samples as the target class $\tau$. However, as shown in Eq. \eqref{eq3} , generating the universal trigger requires $M$ samples with labels $1-\tau$, which is unattainable for the adversary, as it lacks knowledge of any sample label information.

%Although Eq. \eqref{eq3} does not necessitate modification of the target label, in order to carry out a class-specific backdoor attack, the adversary still needs to obtain $m$ poisoned samples labeled $1-\tau$. For instance, if the adversary aims to introduce a backdoor in class 1, it must optimize the universal trigger based on the ascending direction of the negative samples, and the reverse also applies for class 0. However, owing to the split structure of the VFL architecture, the adversary is unable to be aware of the label information for any sample.

%In addition, to accomplish an effective and stealthy backdoor attack, the universal trigger must be class-specific, implying that a limited number of target labels need to be acquired in advance by the adversary. These target samples are utilized to optimize the 

To overcome this limitation, UAB exploits the class skew property inherent in binary classification tasks, enabling it to infer $M$ samples with pseudo labels $1-\tau$ with a certain degree of confidence. As discussed in \cite{li2022label}, the gradients $g=\frac{\partial \mathcal{L}}{\partial H_{i}^A}$ transmitted by the top model of the active party exhibit a significant distribution gap between the negative and positive categories, which can be expressed as:

\begin{equation}\label{eq5}
g=({p}_1-y_i) \cdot \nabla_a G(a)_{a={f_A}(x_i^A)}
\end{equation}

\noindent
where ${p}_1-y_i$ represents the gradients of the binary cross-entropy loss with respect to the logit $z$, and ${p}_1=1 /(1+\exp (-z))$ is the predicted probability of the positive class. $\nabla_a G(a)_{a={f_A}(x_i^A)}$ denotes the gradients of the logit with respect to the embedded features of the adversary.

In real-world scenarios, such as credit loans or disease predictions, the VFL model is typically biased toward negative samples, which constitute the majority of the data. This implies that $p_1$ usually tends towards 0. As a result, the term $\left|{p}_1-y_i\right|$ exhibits larger values for the positive class. Additionally, the term $\|\nabla_a G(a){a={f_A}(x_i^A)}\|_2$ displays a similar order of magnitude for both negative and positive categories, as it is not associated with the labels. Therefore, $\|g\|_2$ can be effectively utilized to infer pseudo labels. 

It should be noted, however, that the magnitude of $\|g\|_2$ focuses on the relative difference between the positive and negative categories and does not provide an accurate prediction of the label for each sample. Therefore, we sort the $\|g\|_2$ values of the samples in each batch and select the portion of samples with the largest values as the pseudo positive samples, and the portion with the smallest values as the pseudo negative samples. 
%In other words, the poisoned set $m$ comprises a fixed number of samples in each training batch, specifically those with the maximum or minimum $|g|_2$ value.

\begin{algorithm}[!t]
 	\caption{ Training for universal trigger}
 	\label{alg:UAP}
 	\begin{algorithmic}[1]
 		\REQUIRE~~\\ Pre-trained model $F(\cdot)$ parameterized by $\Theta^*$, dataset $\mathcal{D}$, target class $\tau$, number of non-target samples $m$ \\
 		 
 		\ENSURE~~\\ Universal  trigger $\delta$
 		%{ {\STATE \bf repeat} Collect newly observed QoS data;}{
        \STATE \textbf{ initialize:} $\delta \leftarrow 0$ \\ 
        \FOR{each batch $B \in \mathcal{D}$ }
        \IF{$\tau$ ==0}
        \STATE $ ind=\arg TopK(|g|_2, m)$  \\
        \ELSE
        \STATE $ ind=\arg TopK(|g|_2, m, largest=Flase)$ \\
        \ENDIF
        \STATE computes $\nabla_{\delta} \mathcal{L}$ with Eq. \eqref{eq4} \\
        \STATE updates $\delta \leftarrow \delta + \alpha \operatorname{sgn}( \nabla_{\delta} \mathcal{L}[ind])$ \\
        \STATE projects $\delta$ to $\|\delta\|_{\infty} \leq \epsilon$ 
        \ENDFOR
 		{\STATE \bf Output $\delta$}
 	\end{algorithmic}
 \end{algorithm}

The training process of the universal trigger is detailed in Algorithm \ref{alg:UAP}. Specifically, for each training batch, the adversary selects $m$ samples with either the smallest or largest values of $\|g\|_2$, depending on the target class $\tau$ (lines 3-7). If the target class is 0, the $m$ samples with the largest $\|g\|_2$ values are assigned a pseudo label of 1. Conversely, if the target class is 1, the $m$ samples with the smallest $\|g\|_2$ values are assigned a pseudo label of 0. Using these $m$ samples with pseudo labels, the universal trigger is optimized in the ascending direction of the gradients $ {\nabla_{\delta} \mathcal{L}}$ and projected in $\ell_{\infty}$-norm \cite{DBLP:journals/corr/Ruder16} (line 8-10). The $M$ non-target samples consists of $m$ samples in each batch.

\subsection{Clean-label backdoor injection}

The training procedure of the universal trigger, as outlined in Algorithm \ref{alg:UAP}, relies on a pre-trained model $F(\cdot)$. This implies that the process is carried out during the inference stage of machine learning. However, this strategy is ill-suited for the VFL architecture, as the gradients $g$ in Eq. \eqref{eq5} are accessible exclusively during the VFL training process. As a result, we adopt sub-optimal model parameters as the pre-trained model $\Theta^*$ that can be obtained in the intermediate training stage in VFL. The training objective can be expressed as:

\begin{equation}\label{eq6}
  F(\Theta^*, \delta) \in \arg \min _{\Theta} \frac{1}{N} \sum_{i=1}^{N}  \ell \left(G\left(H_{i}^1, \cdots, \hat{H}_{i}^{A} ,\cdots, H_{i}^K \right), y_{i}\right)
\end{equation}

\noindent
where $\hat{H}_{i}^{A} = f_{A}(\theta_{A},x_{i}^{A}+\delta)$ denotes the poisoned embedded features of the adversary. $\Theta^*$ denotes the sub-optimal model parameters in the intermediate training stage of VFL. 

From Eq. \eqref{eq6}, it is evident that the universal trigger has an impact on the intermediate model. This is because the generation of the universal trigger is based on the gradients $g$ passed back from the active party. In order to obtain the intermediate gradients for the poisoned samples, the poisoned embedded features containing the trigger must be transmitted to the active party first. The active party will then pass back the gradients $g$ after updating the top model parameters. In addition, as the universal trigger is updated using small batches of training data, it also contaminates the bottom model parameters of the adversary during the iterative training process. In fact, Eq. \eqref{eq6} represents the process of clean-label backdoor injection into the VFL framework.

\subsection{Optimization of the UAB attack}\label{sec:4.2}

In summary, considering Eq. \eqref{eq3} and Eq. \eqref{eq6}, the UAB training in the VFL setting can be formulated as follows:

\begin{equation}\label{eq7}
\begin{aligned}
&  \max _{ \|\delta\|_p \leq \epsilon }  \frac{1}{M} \sum_{i=1}^{M} \mathcal{L}  (F(x_{i}, \Theta(\delta)), 1-\tau)\\
 \text { s.t. } & \Theta(\delta) \in \arg \min _{\Theta} \frac{1}{N} \sum_{i=1}^{N}  \mathcal{L} \left(F(x_i+\delta, \Theta) , y_{i}\right)
\end{aligned}
\end{equation}

\noindent
where the UAB attack is a standard bi-level problem in VFL, which can be effectively solved by alternately optimizing the upper-level and lower-level sub-problems\cite{li2022untargeted}.

%The upper-level sub-problem is the training of the universal trigger, as illustrated in the top of Eq. \eqref{eq7}. Given the current model parameters $\Theta(\delta)$, the optimization of the universal trigger follows Algorithm \ref{alg:UAP}. The lower-level sub-problem is the VFL training in the constraint at the bottom of Eq. \eqref{eq7}. With the updated universal trigger, the model parameters are optimized following Algorithm \ref{alg:VFL_baseline}. 

%Secondly, the training of UAB attack is a standard bi-level optimization, which can be effectively solved by alternatively optimizing the upper-level and lower-level sub-problems. More specifically, the universal trigger $\delta$ is computed based on the model parameters $\Theta$ trained in the previous iteration. At the same time, Then the universal trigger is utilized for the next backdoor iteration.

%For each universal adversarial attack, one clean-backdoor attack in Eq. \eqref{eq6} is followed. To mitigate the performance degradation of the main task in VFL, the UAB attack are distributed throughout the training phase, with one step attack launched every certain number of iterations. By the way, since UAB is injected into the training process of VFL, once the attack occurs in a certain iteration, backdoor and universal adversarial attacks are performed simultaneously. But the backdoor attack is based on UAP obtained from the previous attack round, and the universal adversarial attack generates a trigger to serve the next attack round.

Due to the adversary's lack of knowledge about label information and the fact that visual concealment is not necessary under the VFL split architecture, the adversary uses all local training data as poisoned samples. However, large-scale poisoned data can impair the performance of the main VFL task, potentially raising suspicion among other parties, particularly the active party. Moreover, the bi-level optimization problem in Eq. \eqref{eq7} can have an effect similar to adversarial training, which may also result in performance degradation for the backdoor task. To address these issues, UAB employs a one-time attack strategy. In particular, the UAB attack is executed at a specific iteration during the training process. In this backdoor iteration, Eq. \eqref{eq7} alternately optimizes the upper-level and lower-level sub-problems across multiple batches.

The training procedure of the UAB attack within the VFL framework is outlined in Algorithm \ref{alg:UAB}. If the attack iteration is not reached, the framework follows the VFL baseline training process. Specifically, in each mini-batch $B$ of every iteration, each party computes embedded features $H_{i}^{k}$ using its respective bottom model $f_k$ (lines 7-9). Subsequently, the active party aggregates embedded features $H_{i}$ and computes the objective function as per Eq. \eqref{eq1}. The active party then updates the top model using $\frac{\partial \mathcal{L}}{ \partial \theta_{Top}}$ and sends the gradients $\frac{\partial \mathcal{L}}{\partial H_{i}}$ to all parties (lines 10-12). Lastly, based on the received gradients $\frac{\partial \mathcal{L}}{\partial H_{i}}$, each party calculates the gradients of the loss function with respect to the bottom model parameters $\nabla_{\theta_{k}} \mathcal{L}$ and proceeds to update the bottom model parameters (lines 24-27).

Upon reaching the attack iteration $r$, the UAB attack is introduced into the VFL framework, which involves both the generation of the universal trigger and the injection of the clean-label backdoor. Given that the generation of the universal trigger depends on the gradients $g$ in Eq. \eqref{eq5}, an initialized clean-label backdoor must be established. Specifically, the adversary poisons its inputs by incorporating an initialized universal trigger $\delta$ (lines 4-6). The poisoned inputs are then merged with the VFL baseline training to inject a backdoor. Leveraging the gradients $g$ sent from the active party, the adversary selects $m$ samples labeled as $1-\tau$ and updates the universal trigger with these samples (lines 13-23). The updated universal trigger is then employed for the next clean-label backdoor injection (lines 4-6).

\begin{algorithm}[!t]
 	\caption{A VFL framework with UAB attack.}
 	\label{alg:UAB}
 	\begin{algorithmic}[1]
 		\REQUIRE~~\\ Dataset $\mathcal{D}= \{{x_{i},y_{i}}\}_{i=1}^{N}$; attack iteration $r$, target class $\tau$, number of non-target samples $m$ \\
 		\ENSURE~~\\ Model parameters  $\theta_{1},\theta_{2},\cdots, \theta_{K}, \theta_{Top}$ ;\\
        Universal trigger $\delta$ \\
 		\STATE  initialize: $\delta \leftarrow 0$ \\
        \FOR{each iteration $j=1,2,\cdots$}
            \FOR{ each batch $B$ in $\mathcal{D}$}
                %\STATE /* Clean-label backdoor injection  */ 
                \IF{$j  ==  r $:}
                \STATE \textbf{Adversary} $A$: updates $\{x_i^A = x_i^A+ \delta\}_{i \in B}$
                \ENDIF
                \FOR{each party $k=1,2,\dots, K$ in parallel}
                \STATE $k$ computes embedded features $\{H_{i}^{k}\}_{i \in B}$ according to its bottom model $f_k$;
                \ENDFOR
                \STATE  \textbf{Active party}:
                \STATE  computes Eq. \eqref{eq1}, and then updates $\theta_{Top}$ with $\frac{\partial \mathcal{L}}{ \partial \theta_{Top}}$;
                \STATE  sends $\{\frac{\partial \mathcal{L}}{\partial {H}_{i}}\}_{i \in B}$ to all parties; 
                %\STATE /* Universal trigger generation  */ 
                \IF{$j  ==  r $:}
                \STATE \textbf{Adversary} $A$: 
                \IF{$\tau$ ==0}
                \STATE  $ ind=\arg TopK(|g|_2, m)$  \\
                \ELSE
                \STATE  $ ind=\arg TopK(|g|_2, m, largest=Flase)$ 
                \ENDIF
                %\STATE $A$ computes $g$ according to Eq. \eqref{eq9}
                \STATE computes $\nabla_{\delta} \mathcal{L}$ with Eq. \eqref{eq4} \\
                \STATE updates $\delta \leftarrow \delta + \alpha \operatorname{sgn}( \nabla_{\delta} \mathcal{L}[ind])$ \\
                \STATE projects $\delta$ to $\|\delta\|_{\infty} \leq \epsilon$
                \ENDIF
                \FOR{each party $k=1,2,\dots, K$ in parallel}
                \STATE $k$ computes $\nabla_{\theta_{k}} \mathcal{L}=\{\frac{\partial\mathcal{L}}{\partial H_{i} }\frac{\partial H_{i}^{k}}{\partial \theta_{k}}\}_{i \in B}$; 
                \STATE $k$ updates model parameters $\theta_{k}$;
                \ENDFOR 
        \ENDFOR
        \ENDFOR
 		{\STATE  \bf Output  $\theta_{1},\theta_{2},\cdots, \theta_{K}, \theta_{Top}$; $\delta$}
 	\end{algorithmic}
 \end{algorithm}

\subsection{Evaluation metrics for the UAB attack}

%For the BVFL framework, the main task of VFL is the binary classification and the backdoor task is to train the universal trigger that can misclassify the positive samples into negative ones. 

In the context of the UAB attack within the VFL framework, the main task of VFL is binary classification, while the backdoor task aims to inject a backdoor into VFL. This backdoor allows for the misclassification of source class samples into target class when activated. Owing to the common occurrence of class imbalance in binary VFL classification tasks, particularly in finance and healthcare domains, traditional accuracy evaluation metrics may exhibit bias towards negative samples. This can result in an inadequate assessment of both classification and backdoor task performance. To address this issue, we design the metrics for the UAB attack.

For the main task, we adopt AUC and F1 score to evaluate the performance of the VFL main task \cite{277244,li2022label}. Moreover, considering the need for concealment and effectiveness of the backdoor task, we introduce evaluation metrics that include ASR \cite{li2022backdoor} and ARR. ASR measures the proportion of non-target class samples that are successfully misclassified due to the UAB attack. ARR, on the other hand, evaluates the proportion of target class samples that remain correctly classified under the UAB attack. The metrics are as follows:

\begin{equation}\label{eq10}
\text{ASR} = \frac{\text{Number of successful attacks}}{\text{Total number of non-targeted samples}} \times 100\%
\end{equation}

\begin{equation}\label{eq11}
\text{ARR} = \frac{\text{Number of unsuccessful attacks}}{\text{Total number of targeted samples}} \times 100\%
\end{equation}

\noindent
where ASR and ARR are both constrained between 0 and 100\%, respectively. A higher ASR indicates a more successful UAB attack, while a higher ARR signifies better robustness of target class samples. The ideal backdoor performance aims to misclassify target samples while preserving the correct classification of non-target samples under the UAB attack, i.e., $ASR=100\%$ and $ARR=100\%$.

\section{Experiments}\label{sec:5}

In this section, we design and implement comprehensive experiments to evaluate the effectiveness of our proposed UAB attack. Firstly, we demonstrate the details of experiment setup. Then we propose a VFL backdoor baseline to evaluate the performance of our UAB approach. Based on the backdoor baseline, we conduct a comparison with state-of-the-art backdoor attack methods, highlighting the superiority of the UAB attack. In addition, we analyze the performance of the UAB attack in VFL. Furthermore, we perform ablation experiments on the number of the pseudo labels and attack iteration of UAB, which validates the success of the targeted backdoor attack. Finally, we evaluated the robustness of UAB under state-of-the-art defense methods.

\subsection{Experiment Setup}

The UAB attack was evaluated on the public Zhongyuan and LendingClub datasets\cite{DBLP:journals/corr/abs-2007-13518,DBLP:journals/jsa/ChenDLWH22}. The Zhongyuan dataset is provided by the Zhongyuan Bank of China for the CCF Big and Computing Intelligence Contest\footnote{https://www.datafountain.cn/competitions/530/datasets}. The LendingClub dataset is from an online peer-to-peer lending company that contains credit loans from 2007-2018\footnote{https://www.kaggle.com/datasets/wordsforthewise/lending-club}. Given that financial tabular data is a commonly used dataset in VFL applications, evaluating the UAB attack on datasets such as LendingClub and Zhongyuan provides valuable insight into the potential threat it poses to real-world VFL applications. While UAB is applicable for multiple parties in VFL, we evaluated it with two parties for simplicity. For the sake of distinction, we call them active party and the adversary.  

In the Zhongyuan dataset, we chose 10000 personal public loan records for our experiments, where 8317 samples are negative (fully paid) and 1683 are positive (defaulting applicants). We randomly divided 9000 samples as training data, and the remaining 1000 samples are used for testing. To ensure the fairness of our experiments, we randomly designated 19 of the 37 features as local data for the adversary, with the remaining 18 features and labels assigned to the active party.

For the LendingClub dataset, we followed the pre-processing setting in \cite{DBLP:journals/corr/abs-2007-13518}. The dataset from 2018 is divided into a training set and a testing set with 86291 and 21573 samples, respectively. Out of the total samples, 7596 are positive and 100268 are negative. The 83 features are randomly divided into two parties in VFL, with 41 features and labels on the active party and 42 features included on the adversary. 

The parties are based on the multilayer perceptron model, which consists of three linear layers, the first two of which are followed by a LeakyRelu activation layer. The model structure applies to the active party and the adversary respectively. The adversary acts as a feature extractor including two linear layers with LeakyRelu activation. For the active party, it consists of two parts: the top model serves as a classifier that contains a linear layer with Sigmoid activation and the bottom model is the same as the adversary. The UAB attack in a VFL framework is trained by an SGD optimizer with a learning rate of 0.001, and 100 iterations. The batch sizes for the Zhongyuan and LendingClub datasets are set to 256 and 1024, respectively. For the LendingClub dataset, the hyperparameters $\alpha$ and $\epsilon$ are 0.08 and 10, respectively; on the Zhongyuan dataset, they are 0.2 and 10, respectively.

We conducted the experiments on a machine with the following specifications: A workstation equipped with Intel(R) Xeon(R) Gold 5218 CPU @ 2.30GHz, 64GB RAM, and two NVIDIA Tesla T4 GPU cards. 

\subsection{Attack Baseline}

One of the most significant concerns with VFL backdoor attacks is the adversary's inability to access or manipulate labels in the active party, which restricts the efficacy of traditional backdoor learning approaches\cite{DBLP:journals/corr/abs-1712-05526}. Although there are a few studies available on VFL backdoor attacks, they are predominantly based on assumptions of auxiliary knowledge, such as the presence of clean target samples. To evaluate the effectiveness of VFL backdoor attacks, we establish a backdoor baseline (BB) for VFL by assuming that the adversary can manipulate labels within the active party. In the baseline setting, we randomly selected 1\% of the adversary's training features to inject backdoor triggers generated with random Gaussian noise, following the work in \cite{Liu2021}. Simultaneously, the adversary flips the labels of the poisoned samples to match the backdoor target. Specifically, we randomly sampled 862 poisoned samples from the training set of the LendingClub dataset, of which 68 samples were labeled as positive. In the case of the Zhongyuan dataset, a total of 80 training samples were randomly selected as poisoned samples, out of which 16 samples were labeled as positive. While the manipulation of labels in VFL is unattainable, BB can serve as a strong baseline for evaluating the efficacy of VFL backdoor attacks. BB represents a high level of performance that can be achieved through backdoor attacks within VFL framework, enabling researchers to assess the efficacy of novel VFL backdoor attack methods. In Table \ref{tab:1}, we illustrated the performance of backdoor and main tasks for BB. Without sacrificing much of the main task performance, BB achieves an excellent performance in ASR and ARR metrics.

\subsection{Comparison with the state-of-the-art methods}

\begin{table*} 
\caption{Comparison with state-of-the-art methods on the LendingClub and Zhongyuan datasets. The "Backdoor performance" and "Main performance" denote the backdoor and main tasks of UAB respectively. VFL baseline is a VFL comparison benchmark that represents a clean model that has not been attacked by UAB. (ASR $\uparrow$, RBS $\uparrow$, AUC $\uparrow$ and F1 $\uparrow$. Best results are highlighted in bold.)} \label{tab:1}
    \centering
    \begin{adjustbox}{width=\textwidth}
    \begin{tabular}{ccccccc} 
    \toprule
    Dataset & Backdoor Target &  Methods & ASR & ARR  & AUC & F1 \\
     \midrule
    \multirow{10}*{LendingClub}  & \multirow{6}{*}{Target 0}
    & VFL baseline  & $32.15\pm0.86$ & $99.96\pm0.01$  & $96.24\pm0.21$ & $80.44\pm0.75$  \\
    ~ & ~ & FF& $52.85\pm18.64$ & $44.29\pm30.77$ & $96.24\pm0.21$ & $80.44\pm0.75$  \\
     ~ & ~  & Targeted CoPur & $1.65\pm2.28$ & $12.65\pm16.77$ & $96.24\pm0.21$ & $80.44\pm0.75$ \\
     ~ & ~  & LRB  & $53.99\pm18.42$ & $99.98\pm0.03$ & $93.72\pm4.96$ & $72.40\pm11.06$    \\
     \cline{3-7}
     ~ & ~  &  BB   & $85.90\pm6.61$ & $99.95\pm0.10$ & $96.07\pm0.33$ & $79.67\pm1.09$    \\
     ~ & ~  &  UAB   & $\bm{99.49\pm0.50}$ & $\bm{100\pm0}$ & $\bm{96.26\pm0.19}$ & $\bm{80.59\pm0.49}$     \\
     \cline{2-7}
     & \multirow{6}{*}{Target 1}
     & VFL baseline   & $\bm{0.04\pm0.01}$ & $67.85\pm0.86$ & $96.24\pm0.21$ & $\bm{80.44\pm0.75}$  \\
    ~ & ~ & FF  & $55.71\pm30.77$ & $47.15\pm18.64$ & $96.24\pm0.21$ & $80.44\pm0.75$ \\
     ~ & ~  & Targeted CoPur & $87.35\pm16.77$ & $98.35\pm2.28$ & $96.24\pm0.21$ & $80.44\pm0.75$  \\
      ~ & ~  & LRB   & $0.59\pm1.02$ & $61.51\pm8.89$ & $93.48\pm4.08$ & $71.14\pm13.52$    \\
      \cline{3-7}
       ~ & ~  & BB  & $\bm{99.35\pm0.31}$ & $99.87\pm0.09$ & $96.09\pm0.36$ & $79.93\pm1.09$  \\
       ~ & ~  & UAB   & $99.25\pm0.98$ & $\bm{100\pm0}$ & $\bm{96.28\pm0.12}$ & $\bm{80.80\pm0.41}$ \\
      \midrule
    \multirow{10}*{Zhongyuan}  & \multirow{6}{*}{Target 0}
    & VFL baseline  & $66.38\pm5.91$ & $93.55\pm1.34$ &  $85.77\pm1.25$ & $40.32\pm5.30$  \\
    ~ & ~ & FF  & $10.63\pm13.53$ & $4.81\pm3.85$ & $85.77\pm1.25$ & $40.32\pm5.30$ \\
     ~ & ~  & Targeted CoPur  & $3.62\pm2.39$ & $30.41\pm23.04$ & $85.77\pm1.25$ & $40.32\pm5.30$   \\
      ~ & ~  & LRB  & $99.24\pm1.27$ & $99.82\pm0.30$ & $85.08\pm1.58$ & $41.50\pm3.56$   \\
      \cline{3-7}
       ~ & ~  & BB & $96.34\pm3.88$ & $99.48\pm0.75$ & $\bm{85.94\pm1.32}$ & $42.25\pm7.27$   \\
       ~ & ~  & UAB & $\bm{100\pm0}$ & $\bm{100\pm0}$ & $85.73\pm1.17$ & $\bm{42.55\pm4.32}$ \\
     \cline{2-7}
     & \multirow{6}{*}{Target 1}
     & VFL baseline  &  $6.45\pm1.34$ & $33.62\pm5.91$ & $85.77\pm1.25$ & $40.32\pm5.30$  \\
    ~ & ~ & FF  &  $95.19\pm3.95$ & $89.37\pm13.53$ & $85.77\pm1.25$ & $40.32\pm5.30$ \\
     ~ & ~  & Targeted CoPur & $69.59\pm23.04$ & $96.38\pm2.39$ & $85.77\pm1.25$ & $40.32\pm5.30$ \\
     ~ & ~  & LRB  & $30.17\pm14.11$ & $76.49\pm13.42$ & $84.33\pm1.35$ & $33.52\pm9.73$  \\
     \cline{3-7}
     ~ & ~  & BB  & $\bm{96.94\pm1.27}$ & $98.67\pm1.20$ & $\bm{86.10\pm1.33}$ & $\bm{43.41\pm4.30}$  \\
     ~ & ~  & UAB  & $89.69\pm14.91$ & $\bm{99.22\pm0.64}$ & $85.74\pm1.13$  & $42.79\pm4.66$ \\
      \bottomrule
    \end{tabular}
    \end{adjustbox}
\end{table*}

To demonstrate the superiority of the UAB attack, we compared it with state-of-the-art methods on the LendingClub and Zhongyuan datasets. The comparison experiments were conducted based on the latest representative approaches\cite{liu2022vertical}, including the feature-flipping (FF) attack \cite{liucopur}, targeted CoPur \cite{liucopur}, and LRB \cite{Liu2021}. To verify the generality of UAB, we carried out the experiments 5 times independently and report the average results along with standard deviation.

The FF attack \cite{liucopur,liu2022vertical} simply flipped the sign of the adversary's local embedded features with an amplification in the inference of VFL. In accordance with \cite{liucopur}, the amplification was set to a value of 20. Since the feature-flipping attack is injected during the inference stage of VFL, the performance of main task remains consistent with the VFL baseline.

Targeted CoPur \cite{liucopur} proposed a distributed adversarial attack using the PGD method \cite{DBLP:conf/iclr/MadryMSTV18} during the inference stage of VFL. We implemented the PGD attack with a learning rate of 0.1 and 50 iterations, following the approach described in \cite{liucopur}. As targeted CoPur is an inference attack, the performance of the main task remains consistent with the VFL baseline. Despite the requirement for hard-to-obtain label information, the targeted CoPur attack can still serve as a benchmark for evaluating the effectiveness of backdoor attacks in VFL.

LRB\cite{Liu2021} utilized a few target labels to leverage the label replacement attack in VFL. Following the same experimental setup as described in \cite{Liu2021}, we randomly selected 1\% of the adversary's training features to inject backdoor triggers generated with random Gaussian noise. The amplify rate is set to 10 for best performance. The poisoned set settings are identical to BB; however, LRB is unable to modify the labels. 

Table \ref{tab:1} demonstrates that the UAB attack surpasses all other approaches on the LendingClub dataset, yielding the best results. When the target class is negative, the UAB attack demonstrates superior performance in terms of both ASR and ARR metrics (99.49\% and 100\%). The results indicate that the attack has the ability to manipulate nearly all positive samples, causing them to be misidentified as negative. Conversely, negative samples containing the universal trigger continue to be accurately predicted as negative. Furthermore, under the UAB attack, the performance of the main task does not deteriorate compared to the benchmark; in fact, it exhibits a slight improvement. When the target class is positive, the UAB and BB methods exhibit comparable performance in terms of ASR and ARR metrics, with both achieving the highest level of accuracy. This result suggests that UAB can also tamper with almost all triggered negative samples, while maintaining the positive samples unchanged. Additionally, the UAB attack still maintains optimal performance in the main task.

In the case of the Zhongyuan dataset, UAB achieves the highest level of performance compared to other state-of-the-art methods when the backdoor target is negative. In instances where the backdoor target is positive, the FF method has the best performance and the UAB attack exhibits a 5.5\% decrease in ASR. However, the FF method cannot address the case where the target label is negative. Therefore, UAB remains a highly competitive backdoor attack method on the small-scale Zhongyuan dataset. The observed reduction in backdoor performance of UAB could be attributed to the limited classification capacity of VFL on Zhongyuan dataset. Due to Zhongyuan's limited dataset of 10,000, the model has a restricted ability to accurately identify positive samples. The VFL baseline mistakenly classifies 66.83\% of positive samples as negative samples, which has a negative impact on the generation of universal trigger in the UAB attack for positive backdoor target. 

Furthermore, when compared with BB that operates under strong assumptions, the UAB method demonstrates superior or competitive performance without the need to modify any labels, which proves the excellent performance of proposed approach.

\subsection{Performance of the UAB attack in VFL}

\begin{table*} 
\caption{Performance comparison of VFL baseline, Main Task and Backdoor Task on the LendingClub and Zhongyuan datasets.
} \label{tab:2}
    \centering
    \begin{adjustbox}{width=\textwidth}
    \begin{tabular}{ccccccccccc} 
    \toprule
    Dataset & Backdoor Target &  Methods & ASR & ARR & ACC & Precision & Recall & AUC & F1 \\
     \midrule
    \multirow{6}*{LendingClub}  & \multirow{3}{*}{Target 0}
    & VFL baseline  & $32.15\pm0.86$ & $99.96\pm0.01$ & $97.57\pm0.08$ & $99.24\pm0.19$ & $67.85\pm0.86$ & $96.24\pm0.21$ & $80.44\pm0.75$ \\
    ~ & ~ & Main Task  & $32.02\pm0.68$ & $99.96\pm0.01$ & $\bm{97.57\pm0.08}$ & $\bm{99.26\pm0.23}$ & $\bm{67.98\pm0.68}$ & $\bm{96.26\pm0.19}$ & $\bm{80.59\pm0.49}$ \\
     ~ & ~  & Backdoor Task & $\bm{99.49\pm0.5}$  & $\bm{100\pm0}$ & $92.54\pm0.14$ & $26.36\pm20.61$ & $0.51\pm0.50$ & $72.50\pm15.43$ & $0.99\pm0.96$  \\
     \cline{2-10}
     & \multirow{3}{*}{Target 1}
     & VFL baseline  & $0.04\pm0.01$ & $67.85\pm0.86$ & $97.57\pm0.08$ & $99.24\pm0.19$ & $67.85\pm0.86$ & $96.24\pm0.21$ & $80.44\pm0.75$ \\
    ~ & ~ & Main Task  & $0.04\pm0.01$ & $68.25\pm0.60$ & $\bm{97.59\pm0.07}$ & $\bm{99.33\pm0.16}$ & $68.25\pm0.60$ & $\bm{96.28\pm0.12}$ & $\bm{80.80\pm0.41}$ \\
     ~ & ~  & Backdoor Task & $\bm{99.25\pm0.98}$  & $\bm{100\pm0}$ & $8.20\pm0.82$  & $7.56\pm0.16$ & $\bm{100\pm0}$ & $65.52\pm19.06$ & $14.03\pm0.26$  \\
      \midrule
    \multirow{6}*{Zhongyuan}  & \multirow{3}{*}{Target 0}
    & VFL baseline  & $66.38\pm5.91$ & $93.55\pm1.34$ & $83.49\pm1.20$ & $52.18\pm6.1$ & $33.62\pm5.91$ & $\bm{85.77\pm1.25}$ & $40.32\pm5.30$ \\
    ~ & ~ & Main Task  & $63.25\pm5.48$ & $93.21\pm1.02$ & $\bm{83.72\pm0.91}$ & $\bm{52.15\pm4.0}$ & $\bm{36.75\pm5.48}$ & $85.73\pm1.17$ & $\bm{42.52\pm4.32}$ \\
     ~ & ~  & Backdoor Task & $\bm{100\pm0}$  & $\bm{100\pm0}$ & $83.22\pm1.13$ & 0 & 0 & $76.94\pm7.86$ & 0  \\
     \cline{2-10}
     & \multirow{3}{*}{Target 1}
     & VFL baseline  & $6.45\pm1.34$ & $33.62\pm5.91$ & $83.49\pm1.20$ & $52.18\pm6.1$ & $33.62\pm5.91$ & $\bm{85.77\pm1.25}$ & $40.32\pm5.30$ \\
    ~ & ~ & Main Task  & $6.82\pm1.16$ & $36.75\pm6.2$ & $\bm{83.69\pm1.15}$ & $\bm{52.69\pm3.98}$ & $36.75\pm6.20$ & $85.74\pm1.13$ & $\bm{42.79\pm4.66}$ \\
     ~ & ~  & Backdoor Task & $\bm{89.69\pm14.91}$  & $\bm{99.22\pm0.64}$ & $25.26\pm12.21$  & $18.64\pm3.22$ & $\bm{99.22\pm0.64}$ & $76.78\pm9.23$ & $31.22\pm4.42$  \\
      \bottomrule
    \end{tabular}
    \end{adjustbox}
\end{table*}

Table \ref{tab:2} displays the performance of the UAB attack compared to the VFL baseline on both the LendingClub and Zhongyuan datasets.  "Main Task" denotes the classification performance of VFL under the UAB attack and "Backdoor Task" means the poisoned classification performance of UAB on VFL. 

The VFL baseline model exhibits greater precision values on the LendingClub and Zhonggyuan datasets, scoring 99.24\% and 52.18\%, respectively, when compared to the recall metric. These results imply that the model's predictions lean towards negative samples, which can be attributed to the imbalanced nature of the binary task. This imbalanced nature is also reflected in the UAB attack. For example, on the Zhongyuan dataset, the ASR metric of VFL baseline for the positive and negative target categories were 6.45\% and 66.38\%, respectively. Specifically, the VFL baseline misclassifies 66.38\% of positive samples as negative, while only 6.45\% of negative samples are incorrectly identified as positive on the Zhongyuan dataset. This phenomenon suggests that UAB faces varying levels of difficulty in backdooring positive and negative categories. Consequently, to assess the effectiveness of the UAB attack, it is essential to evaluate its performance on both positive and negative categories as backdoor targets. Likewise, there exists a significant discrepancy in the ASR metric for the positive and negative target backdoor categories in the LendingClub dataset for the VFL baseline model, with values of 0.04\% and 32.15\%, respectively. This difference further emphasizes the requirement for a comprehensive evaluation of the UAB attack on both positive and negative categories as backdoor targets.

In addition, by comparing the results of the main task with the VFL baseline, we can observe the process of backdoor injection process of UAB. Following the UAB attack, the metrics associated with the main task present discernible changes in comparison to those of the VFL baseline. On both the LendingClub and Zhongyuan datasets, the AUC and F1 metrics for the main task deviate significantly from the VFL baseline, irrespective of the backdoor target category being 1 or 0. These observed changes in performance serve as an empirical indication of the backdoor injection process linked to the UAB attack, wherein the VFL model parameters are manipulated based on the universal trigger crafted by the adversary.

Furthermore, Table \ref{tab:2} demonstrates the enhancement of the UAB attack on the model's classification ability for category 1. For instance, on the Zhongyuan dataset, there is a 3.13\% improvement in the recall metric for the main task relative to the baseline model for both backdoor target categories 0 and 1. This contrasts with conventional backdoor attacks, which typically weaken the main task performance. This improvement can be attributed to the fact that UAB utilizes the entire training dataset for poisoning, which in turn increases the diversity of the data to a certain extent, leading to an improvement in the model's generalization capabilities.

\subsection{Visualization of the intermediate gradients}

\begin{figure}[!t]
\centerline{\includegraphics[width=0.55\textwidth]{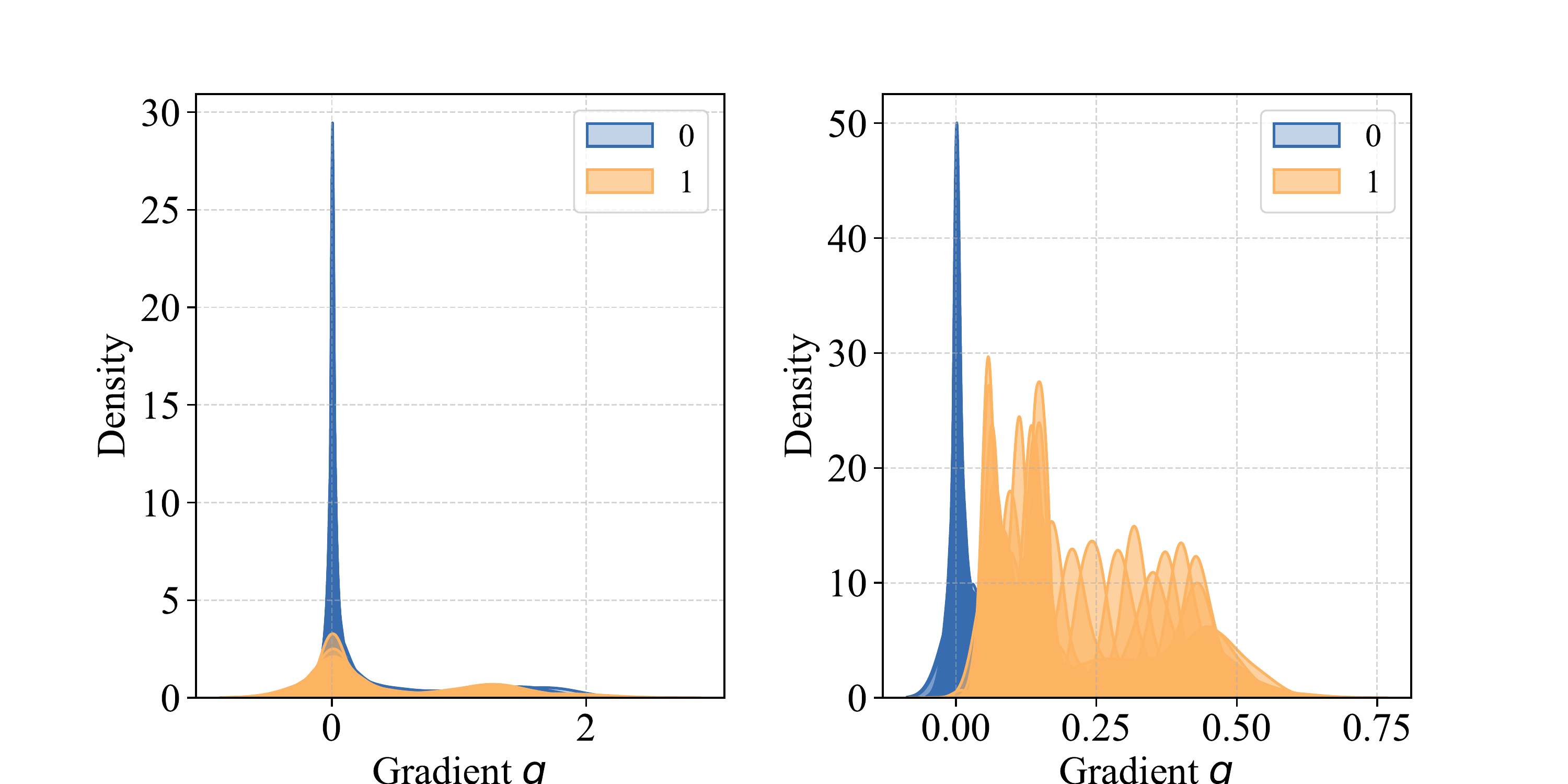}}
\caption{Kernel density estimation of the gradients $g$ for the positive and negative categories on the LendingClub and Zhongyuan training datasets. (left: LendingClub, right: Zhongyuan.)}
\label{fig:3}
\end{figure}

In this part, we presented a qualitative analysis of the universal trigger generation process by utilizing kernel density estimation of the intermediate gradients $g$. The visualizations of these gradients are depicted in Fig. \ref{fig:3}. Both figures display the distribution of the intermediate gradients $g$ for each batch during the UAB attack. The gradients are separated into positive and negative categories based on their corresponding labels.

UAB leverages the class skew property of binary classification tasks to infer that samples with larger gradients are more likely to be positive, while samples with smaller gradients tend to be skewed towards negative samples. The results in Fig. \ref{fig:3} validate the effectiveness of this strategy. 

Specifically, the positive samples in the Zhongyuan and LendingClub datasets actually have larger gradient values compared with the negative ones. For the Zhongyuan dataset, all samples with a gradient value greater than 0.4 are positive; Likewise, all samples with a gradient value greater than 2 are positive for the LendingClub dataset. Therefore, when the target class is 0, a small number of samples with the maximum gradient values can be selected for training the universal trigger, causing the model to recognize samples containing the trigger as class 0. 

Additionally, in comparison to positive samples, negative samples tend to have smaller gradient values that are concentrated around 0. However, it should be noted that some positive samples also have smaller gradient values, particularly in the LendingClub dataset. This is because UAB can only estimate the relative difference between positive and negative inter-class gradients from the class skew property of the dataset. Nevertheless, for large-scale datasets, such as LendingClub, which also exhibit good predictive power for positive class, this can lead to smaller $p_1-y$ values in Eq. \eqref{eq5}. Therefore, for the case where the backdoor target class is 1, more samples with the minimal gradient values should be selected to train the universal trigger.

%In summary, even if only the pseudo label of the sample can be inferred, UAB still enables an efficient backdoor attack method, especially for the case where the target class is 0.

\subsection{Numbers of samples with pseudo labels}

\begin{figure}[!t]
\centerline{\includegraphics[width=0.5\textwidth]{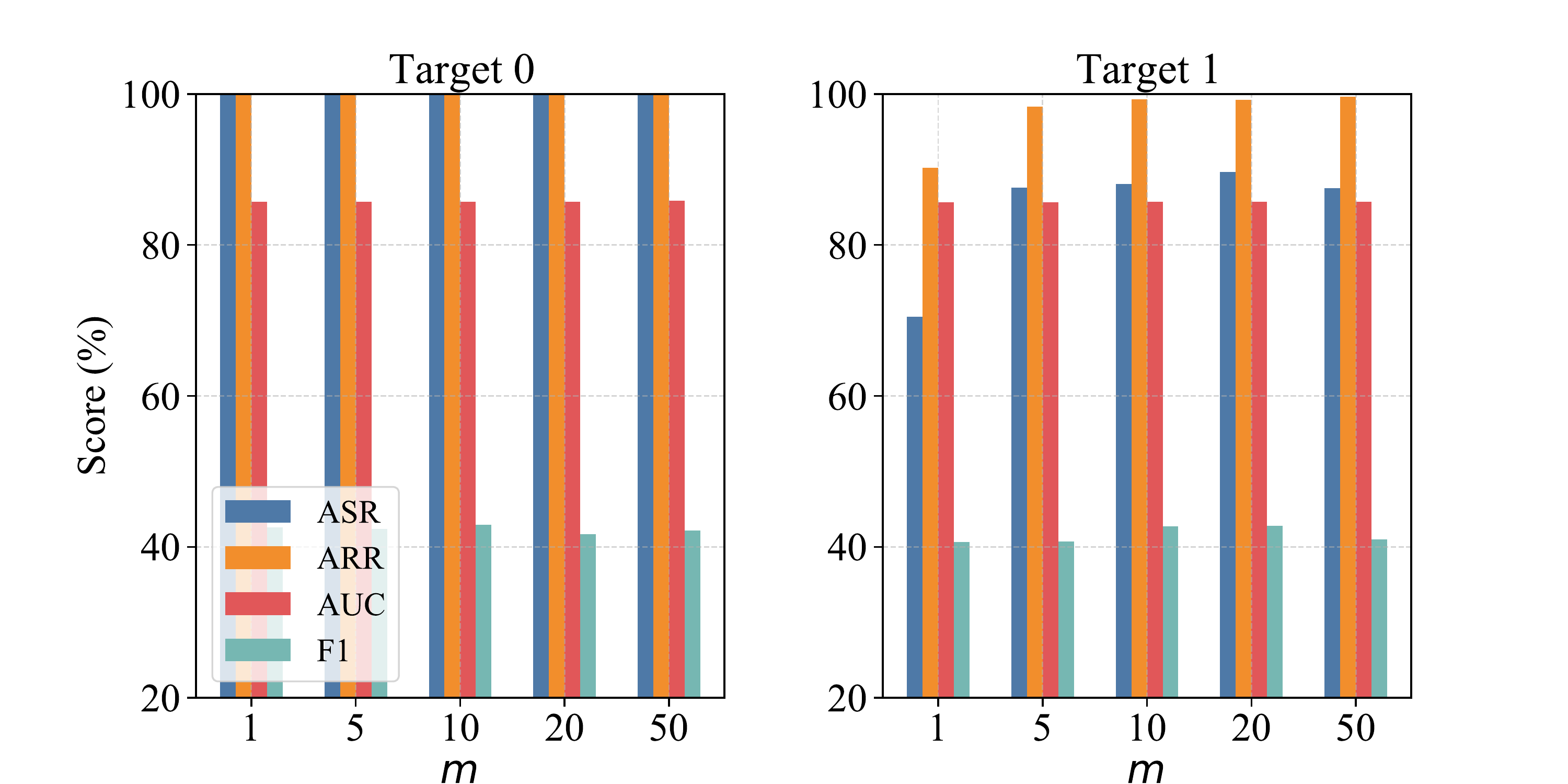}}
\caption{ The performance of UAB with different number of non-target samples $m$ with pseudo labels on the Zhongyuan dataset. }
\label{fig:4}
\end{figure}

\begin{figure}[!t]
\centerline{\includegraphics[width=0.5\textwidth]{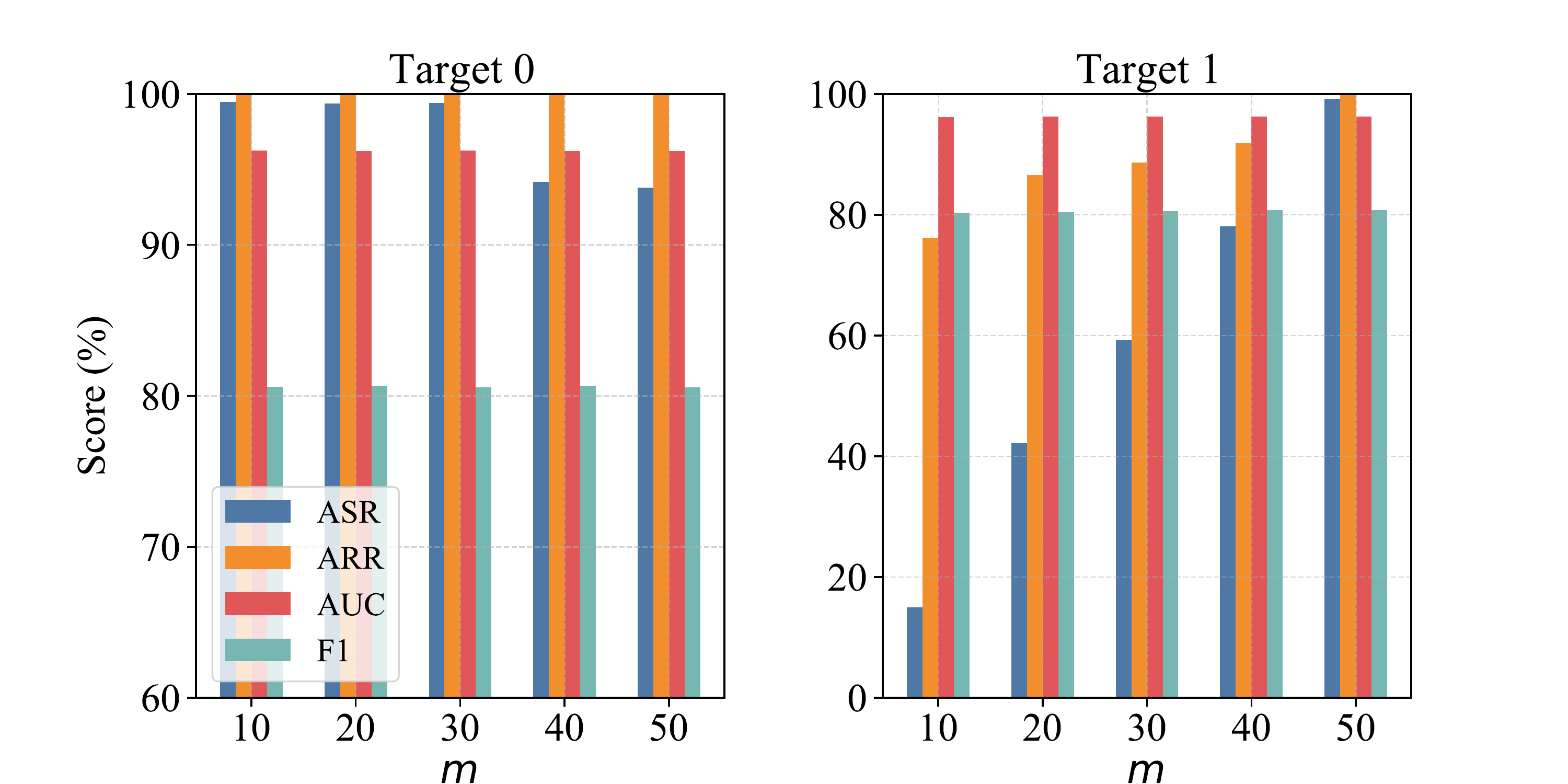}}
\caption{The performance of UAB with different number of non-target samples $m$ on the LendingClub dataset. }
\label{fig:5}
\end{figure}

We conducted experiments to evaluate the effect of the number of  samples with pseudo labels on the performance of the UAB attack. Fig. \ref{fig:4} and \ref{fig:5} depict the performance of UAB on the LendingClub and Zhongyuan datasets when varying the number of $m$ used in the attack process. 

For backdoor target 0, a minimal number of samples with a pseudo label of 1 in each batch of the UAB attack is sufficient to achieve an effective backdoor attack, exhibiting ASR and ARR values close to 100\%. Specifically, only 1 sample is needed for the Zhongyuan dataset, and 10 samples are required for the LendingClub dataset. Furthermore, UAB demonstrates robustness with respect to the number of non-target samples $m$. In particular, When $m=50$, UAB can still maintain a 100\% ASR for the Zhongyuan dataset. For the LendingClub dataset, the attack performance decreases slightly, but it can still maintain an ASR of over 90\%. The performance loss on the LendingClub dataset is attributed to incorrectly predicted pseudo labels. As $m$ increases, some samples with class 0 are also used for the optimization of the universal trigger. This can impact the universal trigger's ability to learn important features of class 0, which in turn affects the overall attack performance. This observation highlights that UAB is a backdoor attack method that does not rely on the availability of ground-truth labels. The adversary leverages the class skew property inherent in binary classification tasks to infer the pseudo labels, and remarkably, UAB still achieves excellent attack performance despite potential inaccuracies in the pseudo labels.

In the case of a backdoor target of 1, UAB requires more samples with pseudo labels 0 to achieve successful backdoor attack. This can be attributed to the inherent model bias towards class 0. Since the universal trigger is optimized based on the model's classification capability, more non-target samples with pseudo labels are necessary to capture important features related to class 1 in order to achieve a robust backdoor attack. Furthermore, ASR of UAB on the LendingClub dataset approaches 100\% when $m=50$, while it only attains an 89\% ASR on the Zhongyuan dataset under the same conditions. This disparity can be attributed to the fact that the UAB attack relies on the classification performance of the model, and the large-scale LendingClub dataset exhibits superior classification performance for class 1 compared to the Zhongyuan dataset.

It is worth noting that in different instances, the performance of the main task is maintained, which suggests that the UAB attack has a relatively minimal impact on the main task performance.

\subsection{Attack iteration of UAB}

\begin{figure}[!t]
\centerline{\includegraphics[width=0.5\textwidth]{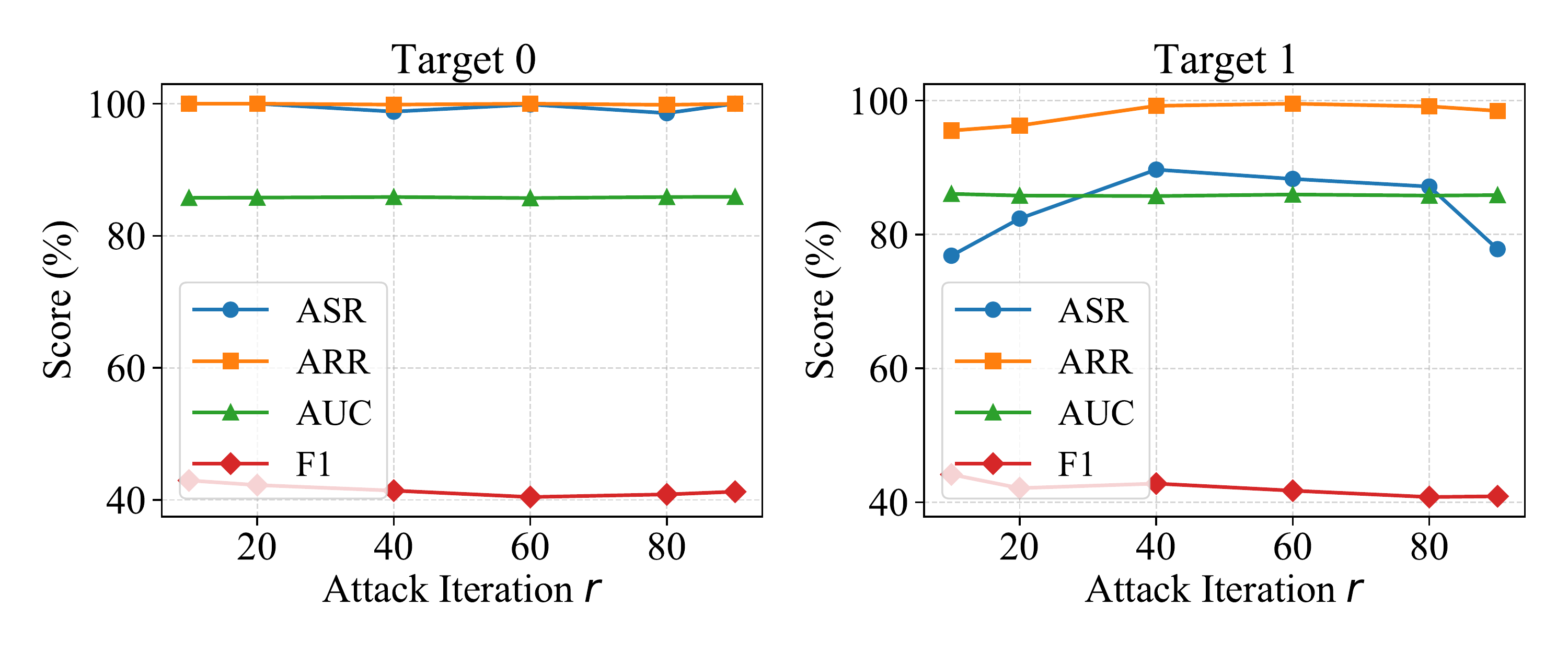}}
\caption{Performance comparison of UAB for different attack iterations on the Zhongyuan dataset.}
\label{fig:6}
\end{figure}

\begin{figure}[!t]
\centerline{\includegraphics[width=0.5\textwidth]{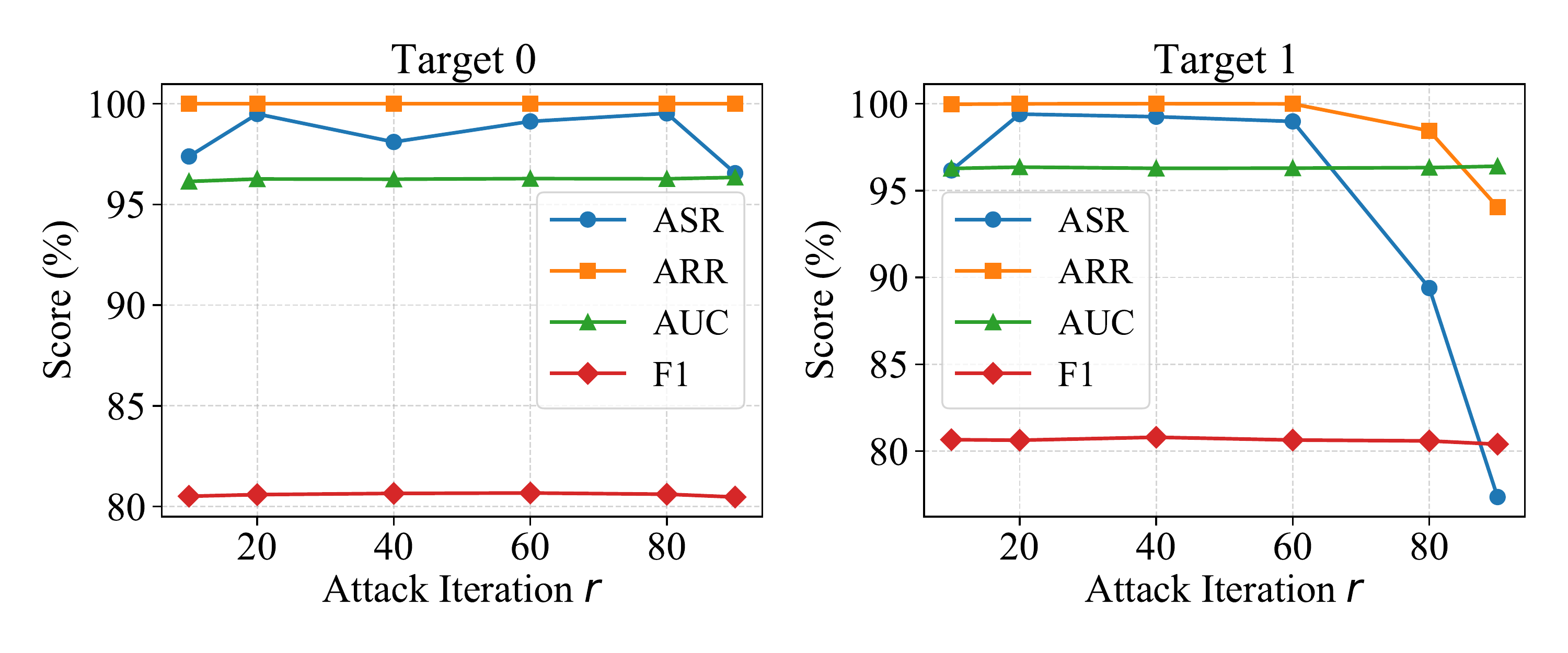}}
\caption{Performance comparison of UAB for different attack iterations on the LendingClub dataset.}
\label{fig:7}
\end{figure}

Algorithm \ref{alg:UAB} illustrates the UAB procedure, which occurs during the training of the VFL model and involves two phases: universal trigger generation and clean-backdoor injection. Backdoor poisoning is achieved through randomized attacks launched at a predetermined number of iterations. This part conducts experimental analysis of the attack iteration of UAB to gain insight into its nature.

Fig. \ref{fig:6} and \ref{fig:7} demonstrate the performance of UAB under different training iterations of VFL. When the backdoor target is 0, excellent backdoor performance can be achieved by launching the UAB attack at various training iterations. For the Zhongyuan and LendingClub datasets, UAB is able to maintain ASR at approximately 100\% and 95\% respectively. In the case of a backdoor target of 1, UAB demonstrates excellent ASR during the training ramp-up and intermediate phases. For instance, on the Zhongyuan dataset, UAB achieves an ASR close to 100\% at 40 iterations, while on the LendingClub dataset, UAB attains an ASR of approximately 100\% at 20 iterations. 

Toward the end of the training process, the performance of UAB experiences varying degrees of degradation, particularly on the LendingClub dataset. In this case, there is a 5\% decrease in ASR for a target category of 0 and a 20\% decrease for a target category of 1, in comparison to the optimal attack performance. This may be because at the end of VFL training, the model tends to converge and the gradient update gradually decreases, which to some extent affects the optimization of the universal trigger in UAB, thus weakening the attack performance\cite{DBLP:conf/iclr/LiuJHCLG020}. Indeed, even at the end of training, UAB maintains a notable ASR of approximately 80\%. This highlights the flexibility and adaptability of the UAB attack, which presents a considerable challenge to the security of the VFL model.

\subsection{Defense evaluation}

\begin{table*} 
\caption{Comparison between different defense methods on the LendingClub and Zhongyuan datasets.} \label{tab:3}
    \centering
    \begin{adjustbox}{width=\textwidth}
    \begin{tabular}{ccccccc} 
    \toprule
    Dataset & Backdoor Target &  Methods & ASR & ARR  & AUC & F1 \\
     \midrule
    \multirow{10}*{LendingClub}  & \multirow{5}{*}{Target 0}
    & VFL baseline  & $32.15\pm0.86$ & $99.96\pm0.01$  & $96.24\pm0.21$ & $80.44\pm0.75$  \\
    ~ & ~ & ISO  & $72.80\pm15.88$ & $\bm{100\pm0}$ & $96.32\pm0.36$ & $79.21\pm1.25$  \\
     ~ & ~  & Max-norm & $\bm{12.57\pm11.22}$  & $57.42\pm41.44$  & $\bm{96.47\pm0.54}$ & $79.36\pm1.75$   \\
     ~ & ~  & Marvell & $45.38\pm31.51$  & $75.08\pm38.68$  & $96.43\pm0.44$ & $\bm{80.59\pm1.01}$   \\
     ~ & ~  & CoPur & $97.43\pm4.51$  & $100\pm0$ & $96.16\pm0.26$ & $79.72\pm1.80$   \\
     \cline{2-7}
     & \multirow{5}{*}{Target 1}
     & VFL baseline   & $\bm{0.04\pm0.01}$ & $67.85\pm0.86$ & $96.24\pm0.21$ & $\bm{80.44\pm0.75}$  \\
    ~ & ~ & ISO   & $55.57\pm38.13$ & $92.52\pm8.67$ & $96.36\pm0.34$ & $79.31\pm1.26$  \\
     ~ & ~  & Max-norm & $26.45\pm26.53$  & $81.59\pm14.16$ & $\bm{96.50\pm0.50}$ & $79.59\pm1.4$   \\
      ~ & ~  & Marvell  & $0.04\pm0.09$  & $47.10\pm18.41$ & $96.29\pm0.45$ & $80.30\pm1.11$   \\
       ~ & ~  & CoPur  & $100\pm0$  & $\bm{100\pm0}$ & $96.28\pm0.60$ & $79.24\pm1.11$   \\
      \midrule
    \multirow{10}*{Zhongyuan}  & \multirow{5}{*}{Target 0}
    & VFL baseline  & $66.38\pm5.91$ & $93.55\pm1.34$ &  $85.77\pm1.25$ & $40.32\pm5.30$  \\
    ~ & ~ & ISO  & $100\pm0$ & $\bm{99.98\pm0.05}$ &  $85.73\pm1.45$ & $38.49\pm5.48$  \\
     ~ & ~  & Max-norm & $\bm{42.83\pm34.30}$  & $72.35\pm31.95$  & $\bm{86.00\pm1.15}$ & $\bm{43.63\pm6.67}$   \\
      ~ & ~  & Marvell & $76.09\pm25.02$  & $94.37\pm6.60$  & $85.66\pm1.50$ & $43.16\pm4.86$   \\
       ~ & ~  & CoPur & $96.86\pm6.27$  & $98.64\pm2.71$  & $78.70\pm3.68$ & $33.14\pm15.48$   \\
     \cline{2-7}
     & \multirow{5}{*}{Target 1}
     & VFL baseline  &  $6.45\pm1.34$ & $33.62\pm5.91$ & $85.77\pm1.25$ & $40.32\pm5.30$  \\
    ~ & ~ & ISO  &  $35.83\pm29.17$ & $66.46\pm28.99$ & $\bm{85.82\pm1.40}$ & $38.51\pm4.78$  \\
     ~ & ~  & Max-norm &  $45.37\pm35.58$  & $78.33\pm17.94$ & $85.78\pm1.09$ & $42.10\pm4.26$  \\
     ~ & ~  & Marvell  & $\bm{28.81\pm32.47}$  & $59.71\pm34.02$ & $85.63\pm1.46$ & $\bm{43.14\pm4.58}$  \\
     ~ & ~  & CoPur  & $86.05\pm27.84$  & $\bm{96.88\pm6.24}$ & $81.15\pm2.97$ & $42.37\pm7.36$  \\
      \bottomrule
    \end{tabular}
    \end{adjustbox}
\end{table*}

To evaluate the robustness of the UAB attack, we analyzed the performance of several representative defense methods. Inspired by \cite{kang2022framework}, we employed four defense methods as benchmarks: Isotropic Gaussian (ISO), Max-norm,  Marvell\cite{li2022label} and CoPur\cite{liucopur}. 

\textbf{ISO} \cite{kang2022framework} is a random perturbation method to mask the actual values of the gradients so that the active party can preserve the labels information. Specifically, ISO adds $iid$ isotropic Gaussian noise to the gradient of each sample, thus confounding the positive and negative labels. In our paper, the ratio is set to 1. 

\textbf{Max-norm} \cite{li2022label} is an improved heuristic version of Gaussian noise. The heuristic Gaussian noise constraints gradients from both magnitude and direction perspectives. The final noise is zero-mean Gaussian with covariance $\sigma=\sqrt{\left\|g_{\max }\right\|_2^2 /\left\|g\right\|_2^2-1}$.

\textbf{Marvell} \cite{li2022label} is an optimization-based noisy protection method. It aims to train the optimal noise distribution for positive and negative samples respectively. Marvell's optimization is to minimize the KL divergence between the perturbed positive and negative gradient distributions\cite{kang2022framework}. The hyperameter $s$ is set to 4 in the experiments.

\textbf{CoPur} \cite{liucopur} is a non-linear robust decomposition method that aims to separate clean features from corrupted features through feature purification. In the experimental setup, we adhere to the methodology presented in \cite{liucopur}.

These methods were chosen to provide a comprehensive evaluation of the model's robustness against the UAB attack. The performance of these defense approaches is summarized in Table \ref{tab:3}.

For an effective defense approach, it is essential to achieve the lowest ASR and the highest ARR, while maintaining the main task performance in comparison to the VFL baseline. Specifically, if the defense method can minimize the misidentification rate of the VFL model while preserving its main task classification performance, it can be inferred that the defense approach effectively protects VFL from the threat posed by the UAB attack. 

On the LendingClub dataset, the Max-Norm method attains the best defense performance with an ASR of 12.57\% when the backdoor target is class 0. This implies that only 12.57\% of the poisoned positive samples will be misclassified as negative. Remarkably, it enhances the classification ability of VFL by 19.58\% for class 1 compared to the VFL baseline. However, the ARR value of Max-Norm drops to 57.42\% relative to the VFL baseline, signifying that 42.58\% of the poisoned negative samples will be misidentified as positive. This suggests that the Max-Norm defense for target class 0 is achieved at the cost of the classification ability of another category. Similarly, for target class 1, the Marvell method obtains the best defense performance with an ASR of 0.04, but concurrently, the ARR value decreases by 20.75\% compared to the VFL baseline.

On the Zhongyuan dataset, for a backdoor target category of 0, the best defense method is Max-Norm, with an ASR of 42.83\%, better than the VFL baseline. However, it sacrifices the classification ability for category 1, with an ARR of 72.35\%, a 21.20\% drop compared to the VFL baseline. When the backdoor target category is 1, Marvell achieves the best defense performance with ASR and ARR of 28.81\% and 59.71\%, respectively. Compared to the VFL baseline, there is still a 22.36\% gap in ASR performance, while ARR exceeds the VFL baseline by 26.29\%. This indicates that although Marvell improves the model's classification ability for category 1, it is still vulnerable to the UAB attack. It is worth noting that both Max-Norm and Marvell methods improve the main task performance, especially in the F1 metric. This may be due to the small size of the Zhongyuan dataset resulting in the significant model bias. The Max-Norm and Marvell methods enhance the equilibrium of the model to some extent through the strategy of positive and negative sample rebalancing.

In summary, even though Max-Norm and Marvell have demonstrated partial defenses, the UAB attack still poses a significant threat to VFL binary classification tasks, which are widely used in practical scenarios.

\section{Conclusion}\label{sec:6}

In this paper, to evaluate the secure in cloud-edge collaboration environments, we present the UAB attack designed to inject a backdoor into binary classification tasks of VFL. The attack consists of two components: universal trigger generation and clean-label backdoor injection. The objective of the universal trigger generation is to train the universal trigger that captures crucial features of the target class. Subsequently, this universal trigger is utilized by the clean-label backdoor injection to poison the model parameters of VFL, leading to misclassifications into the target class. In this study, we opt for a widely used financial scenario to assess the potential security and privacy risks associated with AIoT. Our proposed UAB method outperforms existing state-of-the-art approaches on the LendingClub and Zhongyuan datasets, even attaining an ASR approaching 100\%. Through the assessment of representative defense methods, we establish that UAB indeed presents a significant threat to AIoT systems. This analysis also offers valuable insights that can guide the development of reliable and secure AIoT systems. In the future, we plan to explore the robustness certification against the UAB attack to establish a trusted AIoT framework.

\bibliographystyle{IEEEtran}
\bibliography{UAB}

\end{document}